\documentclass[11pt]{article}

\usepackage{dsfont}

\usepackage{tikz}
\usetikzlibrary{math}
\usepackage{subcaption}
\usepackage[margin=1in]{geometry}

\usepackage{lmodern}
\usepackage{etoolbox}
\usepackage{amsmath}
\usepackage{amssymb}
\usepackage{amsthm}
\usepackage{stmaryrd}
\usepackage[mathic]{mathtools}
\usepackage{bm}

\usepackage{xfrac}

\usepackage{hyperref}

\hypersetup{
    colorlinks=true,
    linkcolor=blue,
    citecolor=blue,
    filecolor=blue,
    urlcolor=blue,
}
\makeatletter
\g@addto@macro\UrlBreaks{\do\/\do\-}
\makeatother

\usepackage[round,authoryear]{natbib}

\usepackage[capitalise]{cleveref}
\newtheorem{theorem}{Theorem}[section]
\newtheorem{lemma}[theorem]{Lemma}
\newtheorem{corollary}[theorem]{Corollary}

\newtheorem{claim}[theorem]{Claim}
\theoremstyle{definition}
\newtheorem{definition}[theorem]{Definition}
\newtheorem{remark}[theorem]{Remark}
\newenvironment{question}
  {%
    \par\noindent\textbf{Question.}\ %
    \begingroup\em  %
  }
  {%
    \endgroup\par
  }

\newcounter{templabel}

\newcommand{\llasttemplabel}{temp:\the\numexpr\value{templabel} - 1\relax}
\newcommand{\lllasttemplabel}{temp:\the\numexpr\value{templabel} - 2\relax}
\newcommand{\llllasttemplabel}{temp:\the\numexpr\value{templabel} - 3\relax}

\newcommand{\placeholderarg}{{\mathchoice{\:}{\:}{\,}{\,}\cdot\mathchoice{\:}{\:}{\,}{\,}}}

\DeclarePairedDelimiter\Par()

\DeclarePairedDelimiter\Ceil\lceil\rceil
\DeclarePairedDelimiter\Floor\lfloor\rfloor

\DeclarePairedDelimiterX\Set[1]\{\}{%
    #1%
}
\DeclarePairedDelimiterXPP\ParWithGiven[1]{}{(}{)}{}{%
    #1%
}
\DeclarePairedDelimiterXPP\ParWithGGiven[1]{}{(}{)}{}{%
    #1%
}
\DeclarePairedDelimiterXPP\BrWithGiven[1]{}{[}{]}{}{%
    #1%
}

\NewDocumentCommand\Prob{ e{_^} }{
    \operatorname{Pr}
    \IfNoValueF{#1}{\sb{#1}}
    \IfNoValueF{#2}{\errmessage{Not supposed to have superscript}\sp{#2}}
    \BrWithGiven
}
\NewDocumentCommand\Ev{ e{_^} }{
    \operatorname{\mathbb{E}}
    \IfNoValueF{#1}{\sb{#1}}
    \IfNoValueF{#2}{\errmessage{Not supposed to have superscript}\sp{#2}}
    \BrWithGiven
}
\NewDocumentCommand\Var{ e{_^} }{
    \operatorname{Var}
    \IfNoValueF{#1}{\sb{#1}}
    \IfNoValueF{#2}{\errmessage{Not supposed to have superscript}\sp{#2}}
    \ParWithGiven
}
\NewDocumentCommand\Cov{ e{_^} }{
    \operatorname{Cov}
    \IfNoValueF{#1}{\sb{#1}}
    \IfNoValueF{#2}{\errmessage{Not supposed to have superscript}\sp{#2}}
    \ParWithGiven
}
\NewDocumentCommand\KL{ e{_^} }{
    \operatorname{KL}
    \IfNoValueF{#1}{\sb{#1}}
    \IfNoValueF{#2}{\errmessage{Not supposed to have superscript}\sp{#2}}
    \ParWithGGiven
}

\NewDocumentCommand{\Fr}{ ssmm }{%
    \IfBooleanTF{#1}{%
        \IfBooleanTF{#2}{%
            #3/(#4)%
        }{%
            #3/#4%
        }%
    }{%
        \mathchoice{%
            \dfrac{#3}{#4}%
        }{%
            \sfrac{#3}{#4}%
        }{%
            \sfrac{#3}{#4}%
        }{%
            \sfrac{#3}{#4}%
        }%
    }%
}
\DeclareMathOperator{\relu}{\varphi}
\DeclareMathOperator{\sign}{sign}

\DeclareMathOperator{\size}{size}

\DeclareMathOperator{\Uniform}{Uniform}

\DeclarePairedDelimiter\abs{\lvert}{\rvert}
\DeclarePairedDelimiterXPP\normzero[1]{}\lVert\rVert{_0}{#1}
\DeclarePairedDelimiterXPP\normone[1]{}\lVert\rVert{_1}{#1}
\DeclarePairedDelimiterXPP\normtwo[1]{}\lVert\rVert{_2}{#1}
\DeclarePairedDelimiterXPP\norminf[1]{}\lVert\rVert{_\infty}{#1}
\DeclarePairedDelimiterXPP\normmax[1]{}\lVert\rVert{_\mathrm{max}}{#1}
\DeclarePairedDelimiterXPP\normspec[1]{}\lVert\rVert{_\mathrm{spectral}}{#1}
\let\norminf\normmax

\NewDocumentCommand\BallClosed{ e{_^} }{
    \operatorname{\mathfrak{B}}_]
    \IfNoValueF{#1}{\errmessage{Not supposed to have subscript}\sb{#1}}
    \IfNoValueF{#2}{\sp{#2}}
    \Par
}
\NewDocumentCommand\BallOpen{ e{_^} }{
    \operatorname{\mathfrak{B}}_)
    \IfNoValueF{#1}{\errmessage{Not supposed to have subscript}\sb{#1}}
    \IfNoValueF{#2}{\sp{#2}}
    \Par
}

\newcommand*{\eps}{\varepsilon}

\let\dd\diff

\DeclarePairedDelimiterX\indicator[1]\llbracket\rrbracket{\,#1\,}   %

\newcommand{\pr}[1]{\text{Pr}\left[#1\right]}
\newcommand{\expect}[1]{\mathbb{E}\left[#1\right]}
\newcommand{\expectwrt}[2]{\mathbb{E}_{#1}\left[#2\right]}

\newcommand{\gaussian}{\cN}
\newcommand{\normal}[1]{\gaussian\!\left(#1\right)}

\newcommand{\st}{\ \middle| \ }

\newcommand{\target}{\vec{z}}
\newcommand{\interval}[1]{I_{#1}}
\newcommand{\multiInterval}[1]{\mathbf{I}_{#1}}
\newcommand{\ones}{\mathbf{1}}
\newcommand{\ind}{\mathds{1}}
\newcommand{\numhits}{Y}
\newcommand{\randSetOne}{\mathrm{S}}
\newcommand{\randSetTwo}{\mathrm{T}}
\newcommand{\eventkernel}{\cE^{\text{kernel}}}
\usepackage{bm}

\makeatletter
\newcommand{\alphacmd@factory}[1]{}
\newcounter{alphacmdcounter}
\newcommand{\GenerateAlphabetCmds}[2]{%
    \renewcommand{\alphacmd@factory}[1]{%
        \expandafter\providecommand\csname #1##1\endcsname{{#2{##1}}}%
    }
    \setcounter{alphacmdcounter}{0}
    \loop
        \stepcounter{alphacmdcounter}
        \edef\alphacmd@ID{\@Alph\c@alphacmdcounter}
        \expandafter\alphacmd@factory\alphacmd@ID
    \ifnum\thealphacmdcounter<26
    \repeat
}

\newcommand{\GenerateAlphabetCmdsLower}[2]{%
    \renewcommand{\alphacmd@factory}[1]{%
        \expandafter\providecommand\csname #1##1\endcsname{{#2{##1}}}%
    }
    \setcounter{alphacmdcounter}{0}
    \loop
        \stepcounter{alphacmdcounter}
        \edef\alphacmd@ID{\@alph\c@alphacmdcounter}
        \expandafter\alphacmd@factory\alphacmd@ID
    \ifnum\thealphacmdcounter<26
    \repeat
}
\makeatother

\GenerateAlphabetCmds{}{\mathbb}

\GenerateAlphabetCmds{c}{\mathcal}

\newcommand*{\StyleRand}{\bm}
\GenerateAlphabetCmds{r}{\StyleRand}
\GenerateAlphabetCmdsLower{r}{\StyleRand}

\newcommand*{\StyleTensor}{\mathnormal}
\GenerateAlphabetCmds{t}{\StyleTensor}
\newcommand*{\StyleRandTensor}[1]{\bm{\StyleTensor{#1}}}
\GenerateAlphabetCmds{rt}{\StyleRandTensor}

\newcommand*{\StyleDist}{\mathcal}
\GenerateAlphabetCmds{d}{\StyleDist}
\newcommand*{\StyleRandDist}[1]{\bm{\StyleDist{#1}}}
\GenerateAlphabetCmds{rd}{\StyleRandDist}

\newcommand*{\StyleSet}{\mathnormal}
\GenerateAlphabetCmds{s}{\StyleSet}
\newcommand*{\StyleRandSet}{\bm}
\GenerateAlphabetCmds{rs}{\StyleRandSet}

\newcommand*{\StyleFamily}{\mathcal}
\GenerateAlphabetCmds{f}{\StyleFamily}
\newcommand*{\StyleRandFamily}[1]{{\bm{\mathcal{#1}}}}
\GenerateAlphabetCmds{rf}{\StyleRandFamily}

\newcommand*{\StyleVec}{\vec}
\GenerateAlphabetCmdsLower{v}{\StyleVec}

\newcommand*{\StyleRandVec}[1]{\StyleRand{\StyleVec{#1}}}
\GenerateAlphabetCmdsLower{rv}{\StyleRandVec}

\definecolor{manim white}{rgb}{1.0, 1.0, 1.0}
\definecolor{manim gray A}{rgb}{0.867, 0.867, 0.867}
\definecolor{manim grey A}{rgb}{0.867, 0.867, 0.867}
\definecolor{manim gray B}{rgb}{0.733, 0.733, 0.733}
\definecolor{manim grey B}{rgb}{0.733, 0.733, 0.733}
\definecolor{manim gray C}{rgb}{0.533, 0.533, 0.533}
\definecolor{manim grey C}{rgb}{0.533, 0.533, 0.533}
\definecolor{manim gray D}{rgb}{0.267, 0.267, 0.267}
\definecolor{manim grey D}{rgb}{0.267, 0.267, 0.267}
\definecolor{manim gray E}{rgb}{0.133, 0.133, 0.133}
\definecolor{manim grey E}{rgb}{0.133, 0.133, 0.133}
\definecolor{manim black}{rgb}{0.0, 0.0, 0.0}
\definecolor{manim lighter gray}{rgb}{0.867, 0.867, 0.867}
\definecolor{manim lighter grey}{rgb}{0.867, 0.867, 0.867}
\definecolor{manim light gray}{rgb}{0.733, 0.733, 0.733}
\definecolor{manim light grey}{rgb}{0.733, 0.733, 0.733}
\definecolor{manim gray}{rgb}{0.533, 0.533, 0.533}
\definecolor{manim grey}{rgb}{0.533, 0.533, 0.533}
\definecolor{manim dark gray}{rgb}{0.267, 0.267, 0.267}
\definecolor{manim dark grey}{rgb}{0.267, 0.267, 0.267}
\definecolor{manim darker gray}{rgb}{0.133, 0.133, 0.133}
\definecolor{manim darker grey}{rgb}{0.133, 0.133, 0.133}
\definecolor{manim blue A}{rgb}{0.78, 0.914, 0.945}
\definecolor{manim blue B}{rgb}{0.612, 0.863, 0.922}
\definecolor{manim blue C}{rgb}{0.345, 0.769, 0.867}
\definecolor{manim blue D}{rgb}{0.161, 0.671, 0.792}
\definecolor{manim blue E}{rgb}{0.137, 0.42, 0.557}
\definecolor{manim blue}{rgb}{0.345, 0.769, 0.867}
\definecolor{manim dark blue}{rgb}{0.137, 0.42, 0.557}
\definecolor{manim teal A}{rgb}{0.675, 0.918, 0.843}
\definecolor{manim teal B}{rgb}{0.463, 0.867, 0.753}
\definecolor{manim teal C}{rgb}{0.361, 0.816, 0.702}
\definecolor{manim teal D}{rgb}{0.333, 0.757, 0.655}
\definecolor{manim teal E}{rgb}{0.286, 0.659, 0.561}
\definecolor{manim teal}{rgb}{0.361, 0.816, 0.702}
\definecolor{manim green A}{rgb}{0.788, 0.886, 0.682}
\definecolor{manim green B}{rgb}{0.651, 0.812, 0.549}
\definecolor{manim green C}{rgb}{0.514, 0.757, 0.404}
\definecolor{manim green D}{rgb}{0.467, 0.69, 0.365}
\definecolor{manim green E}{rgb}{0.412, 0.612, 0.322}
\definecolor{manim green}{rgb}{0.514, 0.757, 0.404}
\definecolor{manim yellow A}{rgb}{1.0, 0.945, 0.714}
\definecolor{manim yellow B}{rgb}{1.0, 0.918, 0.58}
\definecolor{manim yellow C}{rgb}{1.0, 1.0, 0.0}
\definecolor{manim yellow D}{rgb}{0.957, 0.827, 0.271}
\definecolor{manim yellow E}{rgb}{0.91, 0.757, 0.11}
\definecolor{manim yellow}{rgb}{1.0, 1.0, 0.0}
\definecolor{manim gold A}{rgb}{0.969, 0.78, 0.592}
\definecolor{manim gold B}{rgb}{0.976, 0.718, 0.459}
\definecolor{manim gold C}{rgb}{0.941, 0.675, 0.373}
\definecolor{manim gold D}{rgb}{0.882, 0.631, 0.345}
\definecolor{manim gold E}{rgb}{0.78, 0.553, 0.275}
\definecolor{manim gold}{rgb}{0.941, 0.675, 0.373}
\definecolor{manim red A}{rgb}{0.969, 0.631, 0.639}
\definecolor{manim red B}{rgb}{1.0, 0.502, 0.502}
\definecolor{manim red C}{rgb}{0.988, 0.384, 0.333}
\definecolor{manim red D}{rgb}{0.902, 0.353, 0.298}
\definecolor{manim red E}{rgb}{0.812, 0.314, 0.267}
\definecolor{manim red}{rgb}{0.988, 0.384, 0.333}
\definecolor{manim maroon A}{rgb}{0.925, 0.671, 0.757}
\definecolor{manim maroon B}{rgb}{0.925, 0.573, 0.671}
\definecolor{manim maroon C}{rgb}{0.773, 0.373, 0.451}
\definecolor{manim maroon D}{rgb}{0.635, 0.302, 0.38}
\definecolor{manim maroon E}{rgb}{0.58, 0.259, 0.31}
\definecolor{manim maroon}{rgb}{0.773, 0.373, 0.451}
\definecolor{manim purple A}{rgb}{0.792, 0.639, 0.91}
\definecolor{manim purple B}{rgb}{0.694, 0.537, 0.776}
\definecolor{manim purple C}{rgb}{0.604, 0.447, 0.675}
\definecolor{manim purple D}{rgb}{0.443, 0.333, 0.51}
\definecolor{manim purple E}{rgb}{0.392, 0.255, 0.447}
\definecolor{manim purple}{rgb}{0.604, 0.447, 0.675}
\definecolor{manim pink}{rgb}{0.82, 0.278, 0.741}
\definecolor{manim light pink}{rgb}{0.863, 0.459, 0.804}
\definecolor{manim orange}{rgb}{1.0, 0.525, 0.184}
\definecolor{manim light brown}{rgb}{0.804, 0.522, 0.247}
\definecolor{manim dark brown}{rgb}{0.545, 0.271, 0.075}
\definecolor{manim gray brown}{rgb}{0.451, 0.388, 0.341}
\definecolor{manim grey brown}{rgb}{0.451, 0.388, 0.341}

\begin{document}

  \title{Polynomially Overparameterized Convolutional Neural Networks Contain Structured Strong Winning Lottery Tickets}

  \author{
    Arthur da Cunha\\
    Department of Computer Science, Aarhus University\\
    Aarhus, Denmark\\
    \texttt{dac@cs.au.dk}
    \and
    Francesco d'Amore\\
    Department of Computer Science, Gran Sasso Science Institute\\
    L'Aquila, Italy\\
    \texttt{francesco.damore@gssi.it}
    \and
    Emanuele Natale\\
    Team COATI, Universit\'e C\^ote d'Azur, INRIA, CNRS, I3S\\
    Sophia Antipolis, France\\
    \texttt{emanuele.natale@inria.fr}
  }
  \date{}

  \maketitle

\begin{abstract}%
    The Strong Lottery Ticket Hypothesis (SLTH) states that randomly-initialised neural networks likely contain subnetworks that perform well without any training.
    Although unstructured pruning has been extensively studied in this context, its structured counterpart, which can deliver significant computational and memory efficiency gains, has been largely unexplored.
    One of the main reasons for this gap is the limitations of the underlying mathematical tools used in formal analyses of the SLTH.
    In this paper, we overcome these limitations: we leverage recent advances in the multidimensional generalisation of the Random Subset-Sum Problem and obtain a variant that admits the stochastic dependencies that arise when addressing structured pruning in the SLTH.
    We apply this result to prove, for a wide class of random Convolutional Neural Networks, the existence of structured subnetworks that can approximate any sufficiently smaller network.
    
    This result provides the first sub-exponential bound around the SLTH for structured pruning, opening up new avenues for further research on the hypothesis and contributing to the understanding of the role of over-parameterization in deep learning.
\end{abstract}

\section{Introduction}
\label{sec:intro}

Much of the success of deep learning techniques relies on extreme over-parameterization \citep{ZhangBHRV17,ZhangBHRV21,NovakBAPS18,BrutzkusGMS18,DuZPS19,KaplanMHBCCGRWA20}.
While such excess of parameters has allowed neural networks to become the state of the art in many tasks, the associated computational cost limits both the progress of those techniques and their deployment in real-world applications.
This limitation motivated the development of methods for reducing the number of parameters of neural networks; both in the past \citep{Reed93} and in the present \citep{BlalockOFG20,HoeflerABDP21}.

Although pruning methods have traditionally targeted reducing the size of networks for inference purposes, recent works have indicated that they can also be used to reduce parameter counts during training or even at initialisation without sacrificing model accuracy.
In particular, \citet{Frankle19} proposed the \emph{Lottery Ticket Hypothesis (LTH)}, which conjectures that randomly-initialised networks contain sparse subnetworks that can be trained and reach the performance of the fully-trained original network.
Empirical investigations on the LTH \citep{Zhou19,Ramanujan20,WangZXZSZH20} pointed towards an even more striking phenomenon: the existence of subnetworks that perform well without any training.
This conjecture was named the \emph{Strong Lottery Ticket Hypothesis (SLTH)} by \citet{PensiaRNVP20}.

While the SLTH has been proved for many different classes of neural networks (see \cref{sec:related_work}), those works are restricted to unstructured pruning, where the subnetworks are obtained by freely removing individual parameters from the original network.
However, this lack of structure can significantly reduce the gains that sparsity can bring, both in terms of memory and computational efficiency.
The possibility of removing parameters at arbitrary points of the network implies the need to store the indices of the remaining non-zero parameters, which can become a significant overhead with its own research challenges \citep{PoochN73}.
Moreover, the theoretical computational gains of unstructured sparsity can also be difficult to realise in standard hardware, which is optimised for dense operations.
Most notably, the irregularity of the memory access patterns can lead to both data and instruction cache misses, significantly reducing the performance of the pruned network.

\begin{figure}
    \centering
\begingroup
\begin{tikzpicture}[scale=0.25]
    \pgfmathsetseed{123}
    \tikzmath{
        \Nw = 16;
        \Nh = 16;
        \gap = 3;
        \sparsity = 0.4;
        {\fill[manim dark blue] (0, 0) rectangle +(\Nw, \Nh);};
        \w = \Nw;
        for \i in {0, ..., \Nw - 1} {
            if random() < \sparsity then {
                \w = \w - 1;
                {\fill[white] (\i, 0) rectangle +(1, \Nh);};
            };
        };
        \h = \Nh;
        for \i in {0, ..., \Nh - 1} {
            if random() < \sparsity then {
                \h = \h - 1;
                {\fill[white] (0, \i) rectangle +(\Nw, 1);};
            };
        };
        {\draw[step=1, thin, black] (0, 0) grid (\Nw, \Nh);};
        {\draw[->, very thick] (\Nw + \gap, \Nh/2) -- +(\gap, 0);};
    }
    \begin{scope}[shift={(\Nw + 3*\gap, \Nh/2 - \h/2)}]
        \fill[manim dark blue] (0, 0) rectangle +(\w, \h);
        \draw[step=1, thin, black] (0, 0) grid +(\w, \h);
    \end{scope}
\end{tikzpicture}
\endgroup
     \caption[Neuron pruning]{
        Illustration of neuron pruning.
        The left side shows the effect of pruning of neurons in the weight-matrix of a fully-connected layer.
        The rows in white correspond to neurons pruned in the associated layer while the columns in white represent the effect of removing neurons from the previous layers.
        On the right, we allude to the possibility of collapsing the pruned matrix into a smaller but equivalent dense one.
    }
    \label{fig:neuron_pruning}
\end{figure}
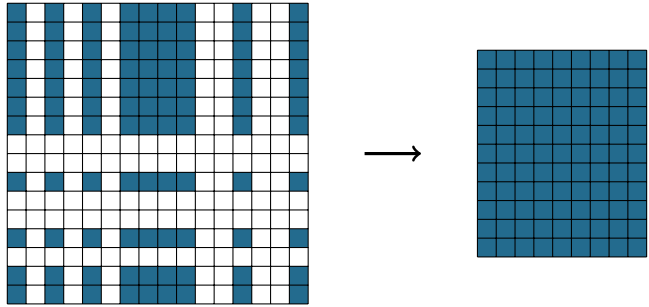

The limitations of parameter-level pruning have motivated extensive research on \emph{structured pruning}, which constrain the sparsity patterns to reduce the complexity of parameter indexation and, more generally, to make the processing of the pruned network more efficient.
A simple example of structured pruning is \emph{neuron pruning} of fully-connected layers: deletions in the weight matrix are constrained to the level of whole rows/columns.
As illustrated by \cref{fig:neuron_pruning}, pruning under this constraint produces a smaller network that is still dense, directly reducing the computational costs without any need for extra memory to store indices.
Similarly, deleting entire filters in Convolutional Neural Networks (CNNs) \citep{PolyakW15} or ``heads'' in attention-based architectures \citep{MichelLN19} also produces direct reductions in computational costs.

\begin{figure}
    \centering
\pgfmathsetseed{12345}

\begin{subfigure}{0.33\textwidth}
    \centering
    \caption*{Unstructured sparsity}
    \begin{tikzpicture}[scale=0.2]
        \foreach \i in {0,...,15} {
            \foreach \j in {0,...,15} {
                \pgfmathparse{rnd > 0.6 ? "manim dark blue" : "white"}
                \fill[\pgfmathresult] (\i, \j) rectangle ++(1, 1);
            }
        }
        \draw[step=1, thin, black] (0, 0) grid (16, 16);
    \end{tikzpicture}
    \caption{No pattern.}
    \label{fig:unstructured}
\end{subfigure}
\begin{subfigure}{0.66\textwidth}
    \centering
    \caption*{Structured sparsity}
    \begin{subfigure}{0.45\textwidth}
        \centering
        \begin{tikzpicture}[scale=0.2,xscale=-1]
            \foreach \i in {0,...,15} {
                \foreach \j in {0,...,15} {
                    \pgfmathparse{mod(\i+\j,3) == 0 ? "manim dark blue" : "white"}
                    \fill[\pgfmathresult] (\i, \j) rectangle ++(1, 1);
                }
            }
            \draw[step=1, thin, black] (0, 0) grid (16, 16);
        \end{tikzpicture}
        \caption{Strided pattern.}
        \label{fig:stride}
    \end{subfigure}
    \begin{subfigure}{0.5\textwidth}
        \centering
        \begin{tikzpicture}[scale=0.2]
            \foreach \i in {0,4,...,15} {
                \foreach \j in {0,4,...,15} {
                    \pgfmathparse{rnd >= 0.6 ? "manim dark blue" : "white"}
                    \fill[\pgfmathresult] (\i, \j) rectangle ++(4, 4);
                }
            }
            \draw[step=1, thin, black] (0, 0) grid (16, 16);
        \end{tikzpicture}
        \caption{Block pattern.}
        \label{fig:block}
    \end{subfigure}
\end{subfigure}
     \caption{Examples of different pruning patterns.}
    \label{fig:pruning_types}
\end{figure}
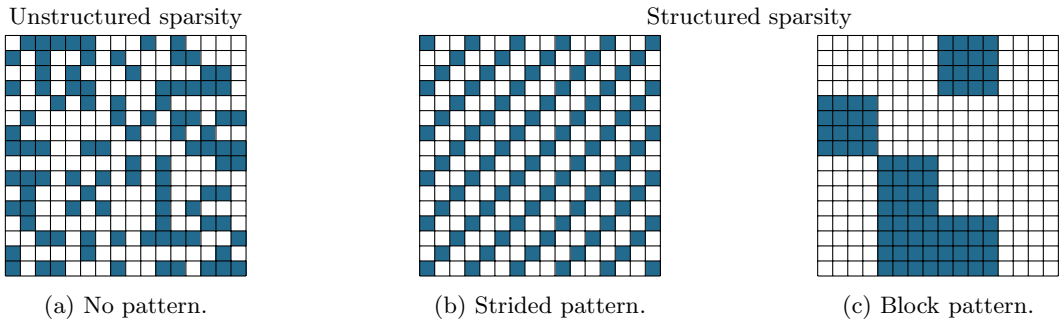

It is important to note that structured pruning is a restriction of unstructured pruning so, theoretically, the former is bound to perform at most as well as the latter.
For example, by deleting whole neurons one can remove about 70\% of the weights in dense networks without significantly affecting accuracy, while through unstructured pruning one can usually reach 95\% sparsity without accuracy loss \citep{AlvarezS16,LiuSZHD19}.
In practice, however, the computational advantage of structured pruning can offset this difference:
a suitably structured sparse network can be more efficient than an even sparser one that lacks structure.
This trade-off between sparsity and actual efficiency has motivated the study of less coarse sparsity patterns since weaker structural constraints such as strided sparsity \citep{AnwarHS17} (\cref{fig:stride}) or block sparsity \citep{Siswanto21} (\cref{fig:block})
are already sufficient to deliver the bulk of the computational gains that structured can offer.

Despite its benefits, structured pruning has received little attention in the context of the SLTH.
In fact, the work that first proved a version of the SLTH, \citet{malachProvingLotteryTicket2020a}, has remained the only one to study this scenario, to the best of our knowledge.
Moreover, it brings a negative result: the authors prove that removing neurons from a randomly-initialised shallow neural network (a single hidden layer) is equivalent to the random features model\footnotemark (e.g., \citep{RahimiR07,RahimiR08}), which cannot efficiently approximate even a single ReLU neuron \citep{YehudaiS19}.
\footnotetext{In the random features model, one can only train the last layer of the network.}
As most proofs around the SLTH involve pruning a shallow random network to build the desired approximations, those results show that the SLTH can be challenging to tackle when restricted to neuron pruning.

In addition, we believe that a factor hindering progress toward structured versions of the SLTH more generally is a limitation of a result underlying almost all the theoretical works on the SLTH: a theorem by \Citeauthor{lueker1998} on the \emph{Random Subset-Sum Problem (RSSP)}.
\begin{theorem}[\citep{lueker1998,da_cunha_revisiting_2022}]\label{thm:lueker}
    Let $X_1, \dots, X_n$ be independent uniform random variables over $[-1, 1]$, and let $\eps \in (0, 1/3)$.
    There exists a universal constant $C > 0$ such that, if $n \ge C\log(1/\eps)$, then, with probability at least $1 - \eps$, for all $z \in [-1, 1]$ there exists $S_z \subseteq [n]$ for which
    \begin{equation*}
        \abs{z - \sum_{i \in S_z} X_i} \le \eps.
    \end{equation*}
\end{theorem}
In general terms, the theorem states that given a rather small number of random variables, there is a high probability that any target value within an interval of interest can be approximated by a sum of a subset of the random variables.
An important remark is that even though \cref{thm:lueker} is stated in terms of uniform random variables, it is not hard to extend it to a wide class of distributions.\footnote{
    Distributions whose probability density function $f$ satisfies $f(x) \ge b$ for all $x \in [-a, a]$, for some constants $a, b > 0$ (see \cite[Corollary 3.3]{lueker1998}).
}

While \cref{thm:lueker} closely matches the setup of the SLTH, it only concerns individual random variables and directly applying it to entire random structures, as needed when considering structured pruning, would require an exponential number of random variables.
The recent works by \citet{borstIntegralityGapBinary2022,becchettiMultidimensionalRandomSubset2022} reduced this gap by proving multidimensional versions of \cref{thm:lueker}.
Still, the intricate manipulation of the network parameters in proofs around the SLTH imposes restrictions that are not covered by those results.

\subsection*{Contributions}
In this work, we overcome those obstacles and prove that random networks in a wide class of CNNs are likely to contain structured subnetworks that approximate any sufficiently smaller CNN.
To the best of our knowledge, our results provide the first sub-exponential bounds around the SLTH for structured pruning of deep neural networks of any kind.
More precisely,
\begin{itemize}
    \item We prove a multidimensional version \cref{thm:lueker} that is robust to some dependencies between coordinates, which is crucial for structured pruning (\cref{thm:mrrs_gaussian_scaled});
    \item We leverage this result and, by combining two types of structured sparsity (block and neuron/filter sparsity), we show that, with high probability and for a wide class of architectures, polynomially over-parameterized random networks can be pruned in a structured manner to approximate any target network (\cref{thm:slth_kernel});
    \item Our results cover CNNs, which generalise fully-connected networks as well as many layer types commonly used in modern architectures, such as pooling and normalisation layers;
    \item Additionally, our pruning scheme focuses on filter pruning, which, like neuron pruning, allows for a direct reduction of the size and computational cost relative to the original CNN.
\end{itemize}
\section{Related Work}
\label{sec:related_work}

\subsection{SLTH}
Put roughly, research on the SLTH revolves around the following question:
\begin{question}
    Given an error margin $\eps > 0$ and a target neural network $f_{\mathrm{target}}$, how large must an architecture $f_\mathrm{random}$ be to ensure that, with high probability on the sampling of its parameters, one can prune $f_\mathrm{random}$ to obtain a subnetwork that approximates $f_\mathrm{target}$ up to output error $\eps$?
\end{question}

\citet{malachProvingLotteryTicket2020a} first proved that, for dense networks with ReLU activations, it was sufficient for $f_\mathrm{random}$ to be twice as deep and polynomially wider than $f_\mathrm{target}$.
\citet{orseauLogarithmicPruningAll2020a} showed that the width overhead could be greatly reduced by sampling parameters from a hyperbolic distribution.
\citet{PensiaRNVP20} improved the original result for a wide class of weight distribution, requiring only a logarithmic width overhead, which the authors proved to be asymptotically optimal.
\citet{daCunhaNV22} generalised those results with optimal bounds to CNNs with non-negative inputs, which \citet{Burkholz22} extended to general inputs and to residual architectures.
\citet{Burkholz22} also reduced the depth overhead to a single extra layer and provided results that include a whole class of activation functions.
\citet{Burkholz23} obtained similar improvements to dense architectures.
\citet{FischerB21} modified many of the previous arguments to take into consideration networks with non-zero biases.
\citet{FerbachTGB22} further generalised previous results on CNNs to general equivariant networks.
\citet{DiffenderferK21} obtained similar SLTH results for binary dense neural networks within polynomial depth and width overhead, which \citet{SreenivasanRSP22} improved to polylogarithmic overhead.
After the conference version of this work appeared, \citet{XiongLK23} studied the SLTH allowing perturbations on the random weights of the initial network, obtaining results with lower overparameterization requirements.
Later, \citet{natale2024} gave guarantees on the size of the lottery ticket (inside the random network) w.r.t.\ the target network.
Finally, \citet{khumar2025quantization} investigated the theoretical guarantees of the SLTH when the random network can only have quantized weigths.

\subsection{Structured pruning}
Works on structured pruning date back to the early days of the field of neural network sparsification with works such as \citet{MozerS88} and \citet{MozerS89}.
Since then, a vast literature has been built around the topic, particularly for the pruning of CNNs.
For a survey of structured pruning in general, we refer the reader to the associated sections of \citep{HoeflerABDP21}, and to \citep{HeX23} for a survey on structured pruning of CNNs.

\subsection{RSSP}
\citet{PensiaRNVP20} introduced the use of theoretical results on the RSSP in arguments around the SLTH, namely \citet[Corollary 3.3]{lueker1998}.
The work by \citet{da_cunha_revisiting_2022} provides an alternative, simpler proof of this result.
\citet{borstIntegralityGapBinary2022} and \citet{becchettiMultidimensionalRandomSubset2022} proved multidimensional versions of the theorem.
\cref{thm:mrrs_gaussian_scaled} diverges from those results in that it supports some dependencies between the entries of random vectors.
\citet{XiongLK23} studied the RSSP allowing perturbations on the random inputs.
\citet{natale2024} gave guarantees on the RSSP when the subsets are constrained to have a specific size.   %
\section{Preliminaries and contribution}

Given $n \in \N$, we denote the set $\{1, \ldots, n\}$ by $[n]$.
The symbol
$*$ represents the discrete convolution operation and
$\odot$ represents the element-wise (Hadamard) product between tensors, and
$\relu$ represents the (element-wise) ReLU activation function.
The notation $\normone*{\placeholderarg}$ refers to the sum of the absolute values of each entry in a tensor.
Similarly, $\normtwo*{\placeholderarg}$ refers to the square root of the sum of the squares of each entry in a tensor.
$\normmax*{\placeholderarg}$ denotes the maximum norm: the maximum among the absolute value of each entry.
Sometimes we represent a tensor \(\tX \in \R^{d_1 \times \ldots \times d_n}\) via the domain of its indices, as in \(\tX = (\tX_{i_1, \ldots, i_n})_{i_1 \in [d_1], \ldots, i_n \in [d_n]}\).
We denote the normal probability distribution with mean \(\mu\) and variance \(\sigma^2\) by \(\normal{\mu, \sigma^2}\).
We write $\rtU \sim \normal{d_1 \times \ldots \times d_n}$ to denote that $\rtU$ is a random tensor of size $d_1 \times \ldots \times d_n$ with independent and identically distributed (i.i.d.) entries, each following $\normal{0, 1}$.
We refer to such random tensors as \emph{normal tensors}.
Finally, we refer to the axis of a 4-D tensor as \emph{rows}, \emph{columns}, \emph{channels}, and \emph{kernels} (a.k.a. filters), in this order.

For the sake of simplicity, we assume CNNs to be of the form $N\colon [-1, 1]^{D \times D \times c_0} \to \R^{D \times D \times c_\ell}$ given by
$$
  N(\tX) = \tK^\ell * \relu(\tK^{\ell-1} * \cdots \relu(\tK^{1} * \tX)),
$$
where $\tK^i \in \R^{\sqrt{d_i} \times \sqrt{d_i} \times c_{i-1} \times c_i}$ for $i \in [\ell]$,
and the convolutions have no bias and are suitably padded with zeros.
Moreover, when the kernels $\tK^{(i)}$ are normal tensors, we say that $N$ is a \emph{random CNN}.

Our main result is the following.
\begin{theorem}[Structured SLTH]
  \label{thm:slth_kernel}
  Let $D, c_{0}, \ell \in \N$, and $\eps \in \R_{>0}$.
  For $i \in [\ell]$, let $d_i, c_i, n_i \in \N$.
  Let $\cF$ be the class of functions from $[-1, 1]^{D \times D \times c_0}$ to $\R^{D \times D \times c_\ell}$ such that, for each $f \in \cF$
  \begin{align}
    f(\tX) = \tK^{(\ell)} * \relu(\tK^{(\ell-1)} * \cdots \relu(\tK^{(1)} * \tX))
    ,
  \end{align}
  where, for $i \in [\ell]$,
  $\tK^{(i)} \in \R^{\sqrt{d_i} \times \sqrt{d_i} \times c_{i-1} \times c_i}$ and
  $\normone*{\tK^{(i)}} \le 1$.
  Let also $N_0\colon [-1, 1]^{D \times D \times c_0} \to \R^{D \times D \times c_\ell}$ be a $2\ell$-layered random CNN given by
  \begin{align}
    N_0(\tX) = \tL^{(2\ell)} * \relu(\tL^{(2\ell-1)} * \cdots \relu(\tL^{(1)} * \tX))
    ,
  \end{align}
  where for $i \in [\ell]$ the kernels $\tL^{(2i-1)}$ and $\tL^{(2i)}$ are normal tensors of shape $1 \times 1 \times c_{i-1} \times 2n_i c_{i-1}$ and $\sqrt{d_{i}} \times \sqrt{d_{i}} \times 2n_i c_{i-1} \times c_i$, respectively.
  Finally, let $\cG$ be the class of subnetworks that can be obtained by pruning contiguous blocks of parameters and removing entire filters from $N_0$.
  There exists a universal constant $C > 0$, such that
  if, for all $i \in [\ell]$,
  \[
    n_i \ge Cd_i^{5}c_{i}^{5}\log^{2}\frac{d_i c_{i}c_{i-1}\ell}{\eps}
    ,
  \]
  then,
  with probability at least $1 - \eps$,
  we have that, for all $f \in \cF$,
  \[
    \sup_{\tX \in [-1, 1]^{D \times D \times c_0}}
    \min_{g \in \cG}
    \normmax*{f(\tX) - g(\tX)}
    \le \eps
    .
  \]
\end{theorem}
The filter removals ensured by \cref{thm:slth_kernel} take place at layers $1, 3, \ldots, 2\ell-1$ and imply the removal of the corresponding channels in the next layer.
The overall modification yields a CNN with kernels $\widetilde{L}^{(1)}, \ldots, \widetilde{L}^{(2\ell)}$ such that, for $i \in [\ell]$, the kernels $\widetilde{L}^{(2i-1)}$ and $\widetilde{L}^{(2i)}$ have shape $1 \times 1 \times c_{i-1} \times 2c_{i-1} m_i$ and $\sqrt{d_i} \times \sqrt{d_i} \times 2c_{i-1} m_i \times c_{i}$, respectively, where $m_i = \sqrt{n_i / (C_1\sqrt{d_i}\log(1/\eps))}$ for a universal constant $C_1$.
Moreover, our proof ensures that the kernels $\widetilde{L}^{(2i-1)}$ can be required to have a specific type of block sparsity: they can be structured as if pruned by $2m_i$-channel-blocked masks, defined as follows.
\begin{definition}[$n$-channel-blocked mask]
  \label{def:channel_blocked}
  Given a positive integer $n$, a binary tensor $S \in \{0, 1\}^{\sqrt{d} \times \sqrt{d} \times c \times cn}$ is called $n$-channel-blocked if and only if
  \[
    S_{i, j, k, \ell} = \begin{cases}
      1 & \text{if } \left\lceil \frac{\ell}{n} \right\rceil = k, \\
      0 & \text{otherwise},
    \end{cases}
  \]
  for all $i, j \in [\sqrt{d}]$, $k \in [c]$, and $\ell \in [cn]$.
\end{definition}

We remark that, from a broader perspective, the central aspect of \cref{thm:slth_kernel} is that the lower bound on the size of the random CNN depends only on the kernel sizes of the CNNs being approximated.

In \cref{subsec:proof_slth_kernel} we discuss the proof of \cref{thm:slth_kernel}.
It requires handling subset-sum problems on multiple random variables at once (random vectors).
Furthermore, the inherent parameter-sharing of CNNs creates a specific type of stochastic dependency between coordinates of the random vectors, which we capture with the following definition.
\begin{definition}[NSN vector]
  \label{def:nsn_vector}
  A $d$-dimensional random vector $Y$ follows
  a \emph{normally-scaled normal} (NSN) distribution if, for each $i \in \left[d\right]$,
  $Y_{i} = Z \cdot Z_{i}$ where $Z, Z_{1}, \ldots, Z_{d}$ are i.i.d.\ random
  variables following a standard normal distribution.
\end{definition}
A key technical contribution of ours is a Multidimensional Random Subset Sum (MRSS) result that supports NSN vectors.
In \cref{subsec:NSN_analysis} we discuss the proof of the next theorem,
which follows a strategy similar to that of \citep{borstIntegralityGapBinary2022,becchettiMultidimensionalRandomSubset2022}.
\begin{theorem}[Normally-scaled MRSS]
  \label{thm:mrrs_gaussian_scaled}
  Let \(0 < \eps < 1/4\), and let $d$ be a positive integer.
  Suppose $X_{1}, \ldots, X_{n}$ are \(n\) $d$-dimensional i.i.d.\ NSN random vectors, and let \(k = n/(6\sqrt{d})\).
  There exist two positive constant \(C, \delta > 0\) such that, if \(n \ge C d^4 \log \frac{d}{\eps}\), for any $\target \in \R^{d}$ with $\normone*{\target} \le \sqrt{k}$, with probability \(\delta\)
  there exists a subset $S \subseteq [n]$ of size $k$ such that $\normmax*{\left(\sum_{i \in S}X_{i}\right)-\target} \le \eps$.
\end{theorem}
While it is possible to naively apply \cref{thm:lueker} to obtain a version of \cref{thm:slth_kernel}, doing so leads to an exponential lower bound on the required number of random vectors.

\subsection{Comparison with conference version}

With respect to the conference version of this work \citep{dacunha2023polynomially}, the main differences are the following.
\begin{itemize}
  \item We improved the theoretical guarantees of the multidimensional RSSP for NSN vectors (\cref{thm:mrrs_gaussian_scaled}).
  More specifically, in \citep{dacunha2023polynomially} the dependency on the dimension \(d\) was \(d^6 \log^2 (d/\varepsilon)\), while in the current version it is \(d^4 \log (d/\varepsilon)\), with \(\varepsilon\) being the approximation error that we allow.
  We achieved this improvement by refining several steps of the proof and exploiting properties of convolutions and \emph{radially-monotone functions} (\cref{app:radial}), the latter of which was not considered in \citep{dacunha2023polynomially}.
  \item Improvements in \cref{thm:mrrs_gaussian_scaled} directly implies improvements in the main SLTH result (\cref{thm:slth_kernel}).
  More specifically, assume the target network has \(\ell\) kernels. 
  In \citep{dacunha2023polynomially}, if a kernel in the target network has size \(\sqrt{d} \times \sqrt{d} \times a \times b\), the required overparameterization was of the order \(d^7 b^6 \log^3 (d ab \ell / \varepsilon)\), while in the current version it is \(d^5 b^5 \log^2 (d ab \ell / \varepsilon)\), with \(\varepsilon\) being the approximation error that we allow.
\end{itemize}
This is a polynomial improvement with respect to the original bounds and a first step towards the optimal bounds for MRSS problem and, in particular, for the structured SLTH.
Our conjecture is that the right bound for the MRSS problem should be closer to an almost-linear dependency in the dimension \(d\).   %
\section{Analysis}
\label{sec:analysis}

In this section, after proving our MRSS result (\cref{thm:mrrs_gaussian_scaled}), we discuss how to use it to obtain our main result on structured pruning (\cref{thm:slth_kernel}).
Full proofs are deferred to \cref{app:supporting-results}.

\subsection{Multidimensional Random Subset Sum for normally-scaled normal vectors\label{subsec:NSN_analysis}}

We first present some notation.%
Given a set $S$ and a positive integer $n$, we denote by $\binom{S}{n}$ the family of all subsets of $S$ containing exactly $n$ elements of $S$.
Given $\eps \in \R_{>0}$,
we define the interval $\interval{\eps}{(\target_i)} = \left[\target_{i}-\eps, \target_{i} + \eps\right]$.
and the multi-interval $\multiInterval{\eps}(\target) = \left[\target-\eps\ones, \target + \eps\ones\right]$, where \(\ones = (1, 1, \ldots, 1) \in \R^d\).
Moreover, for any event \(\cE\), we denote its complementary event by \(\overline{\cE}\).

In this subsection, we give bounds on the multidimensional random subset sum problem, that is, we estimate the probability that a set of \(n\) random vectors contains a subset of any given size \(k\) that sums up to a value that is \(\eps\)-close to a given target.
The following definition formalises this notion.
\begin{definition}[Subset-sum number]
  \label{def:subsetsumnumber}
  Let \(k \in [n]\).
  Given (possibly random) vectors $X_{1}, \ldots, X_{n}$
  and a vector $\target$, we define the \emph{$(\eps, k)$-subset-sum number}
  of $X_{1}, \ldots, X_{n}$ for $\target$ as
  \[
    \numhits_{\target, \eps, k}(X_1, \ldots, X_n)
    = \sum_{S \in \binom{[n]}{k}} \ind_{\cH_{\target, \eps}(S)}
    ,
  \]
  where $\cH_{\target, \eps}(X_1, \ldots, X_n)$ denotes the event $\normmax*{\target - \sum_{i \in S} X_{i}} \le \eps$, and \(\ind\) denotes the indicator function.
  We write simply $\numhits$ and \(\cH_S\) when $X_{1}, \ldots, X_{n}$, $\target$, \(\eps\), and \(k\)
  are clear from the context.
\end{definition}

To prove \cref{thm:mrrs_gaussian_scaled} we use the second moment method to provide a lower bound on the probability that the subset-sum number \(\numhits\) is strictly positive, which implies that at least one subset of the random vectors suitably approximates \(\target\).
Hence, we seek a lower bound on \(\expect{\numhits}^2 / \expect{\numhits^2}\).

Our first lemma allows to infer a lower bound on \(\expect{\numhits}\): it provides a lower bound on the probability that a sum of NSN vectors is \(\eps\)-close to a target vector.

\begin{lemma}[Sum of NSN vectors]
  \label{lemma:prob_one_set}
  Let $k \in \N$, $\eps \in \left(0, \frac{1}{4}\right)$,
  $\target \in \R^{d}$ such that $\normone*{\target} \le \sqrt{k}$ and $k \ge 16$.
  Furthermore, let $X_{1}, \ldots, X_{k}$
  be $d$-dimensional i.i.d.\ NSN random vectors with $d \le k$, and let
  $c_{d} \le \min\left\{\frac{1}{d}, \frac{1}{16}\right\}$.
  It holds that
  \[
    \pr{\sum_{i = 1}^{k}X_{i} \in \multiInterval{\eps}(\target)} \ge \frac{1}{16}\biggl(\frac{2\eps}{\sqrt{2\pi\left(1 + 2\sqrt{c_{d}} + 2c_{d}\right)k}}\biggr)^{d}
    .
  \]
\end{lemma}
\begin{proof}
  By \cref{def:nsn_vector}, the $j$-th entry of each vector
  $X_{i}$ is $\left(X_{i}\right)_{j} = Z_{i} \cdot Z_{i, j}$ where each
  $Z_{i}$ and $Z_{i, j}$ are i.i.d.\ random variables following a standard
  normal distribution.
  Define the event
  \[
    \cE = \left\{k\left(1 - 2\sqrt{c_{d}}\right)
    \le \sum_{i = 1}^{k}Z_{i}^{2} \le k\left(1 + 2\sqrt{c_{d}} + 2c_{d}\right)\right\}
    ,
  \]
  and let $X = \sum_{i = 1}^{k}X_{i}$.
  By the law of total probability, it holds that
  \begin{align*}
    \pr{X \in \multiInterval{\eps}(\target)}
     & \ge \pr{X \in \multiInterval{\eps}(\target) \st \cE} \pr{\cE}
    \\ & = \expectwrt{Z_{1}, \ldots, Z_{k}}{\pr{X \in \multiInterval{\eps}(\target) \st Z_{1}, \ldots, Z_{k}} \st \cE} \pr{\cE}
    .
  \end{align*}
  Conditional on \(Z_{1}, \ldots, Z_{k}\), the \(d\) entries of \(X\) are independent.
  It follows that
  \begin{align}
    & \expectwrt{Z_{1}, \ldots, Z_{k}}{\pr{X \in \multiInterval{\eps}(\target) \st Z_{1}, \ldots, Z_{k}}\st \cE} \pr{\cE} \nonumber
    \\ = \ & \expectwrt{Z_{1}, \ldots, Z_{k}}{\prod_{j = 1}^{d}\pr{\left(X\right)_{j} \in \interval{\eps}{(\target_j)} \st Z_{1}, \ldots, Z_{k}}\st \cE} \pr{\cE} \nonumber
    .
  \end{align}
  Conditional on \(Z_{1}, \ldots, Z_{k}\), we have that \((X)_j \sim \normal{0, \sum_{i = 1}^k Z_i^2}\).
  Hence,
  \begin{align*}
    & \expectwrt{Z_{1}, \ldots, Z_{k}}{\prod_{j = 1}^{d}\pr{\left(X\right)_{j} \in \interval{\eps}{(\target_j)} \st Z_{1}, \ldots, Z_{k}} \st \cE}\pr{\cE}
    \\ \ge \ & \expectwrt{Z_{1}, \ldots, Z_{k}}{\prod_{j = 1}^{d}\left(\frac{2\eps}{\sqrt{2\pi\left(\sum_{i = 1}^{k}Z_{i}^{2}\right)}}\exp\left(-\frac{\left(\left|z_{i}\right| + \eps\right)^{2}}{2\sum_{i = 1}^{k}Z_{i}^{2}}\right)\right)\st \cE} \cdot \pr{\cE}
    .
  \end{align*}
  Notice that the term \({\sum_i Z_i^2}\) is a sum of chi-square random variables, for which there are known concentration bounds (\cref{lemma:chisquared_bound}).
  By definition of \(\cE\) and by applying \cref{lemma:chisquared_bound} to estimate the term \(\pr{\cE}\), we get that
  \begin{align*}
    & \expectwrt{Z_{1}, \ldots, Z_{k}}{\prod_{j = 1}^{d}\left(\frac{2\eps}{\sqrt{2\pi\left(\sum_{i = 1}^{k}Z_{i}^{2}\right)}}\exp\left(-\frac{\left(\left|z_{i}\right| + \eps\right)^{2}}{2\sum_{i = 1}^{k}Z_{i}^{2}}\right)\right)\st \cE} \cdot \pr{\cE}
    \\ \ge \ & \left(\frac{2\eps}{\sqrt{2\pi\left(1 + 2\sqrt{c_{d}} + 2c_{d}\right)k}}\right)^{d}\exp\left(-\frac{\sum_{i}\left|z_{i}\right|^{2} + 2\eps \sum_{i}\left|z_{i}\right| + d\eps^{2}}{2\left(1 - 2\sqrt{c_{d}}\right)k}\right)\pr{\cE}
    \\ = \ & \left(\frac{2\eps}{\sqrt{2\pi\left(1 + 2\sqrt{c_{d}} + 2c_{d}\right)k}}\right)^{d}\exp\left(-\frac{\normtwo*{\target}^{2} + 2\eps \normone*{\target} + d\eps^{2}}{2\left(1 - 2\sqrt{c_{d}}\right)k}\right)\pr{\cE}
    \\ \ge \ & \left(\frac{2\eps}{\sqrt{2\pi\left(1 + 2\sqrt{c_{d}} + 2c_{d}\right)k}}\right)^{d}\exp\left(-\frac{\normtwo*{\target}^{2} + 2\eps\normone*{\target} + d\eps^{2}}{2\left(1 - 2\sqrt{c_{d}}\right)k}\right)\left(1 - 2e^{-c_{d}k}\right)
    .
  \end{align*}
  As \(c_d k \ge 1\) by hypothesis, it holds that \(1 - 2e^{-c_{d}k} \ge 1/4\).
  Then,
  \begin{align}
    & \left(\frac{2\eps}{\sqrt{2\pi\left(1 + 2\sqrt{c_{d}} + 2c_{d}\right)k}}\right)^{d}\exp\left(-\frac{\normtwo*{\target}^{2} + 2\eps\normone*{\target} + d\eps^{2}}{2\left(1 - 2\sqrt{c_{d}}\right)k}\right)\left(1 - 2e^{-c_{d}k}\right) \nonumber
    \\ \ge \ & \frac{1}{4} \left(\frac{2\eps}{\sqrt{2\pi\left(1 + 2\sqrt{c_{d}} + 2c_{d}\right)k}}\right)^{d}\exp\left(-\frac{\normtwo*{\target}^{2} + 2\eps\normone*{\target} + d\eps^{2}}{2\left(1 - 2\sqrt{c_{d}}\right)k}\right) \nonumber
    \\ \ge \ & \frac 14 \left(\frac{2\eps}{\sqrt{2\pi\left(1 + 2\sqrt{c_{d}} + 2c_{d}\right)k}}\right)^{d}\exp\left(-\frac{k + 2\eps\sqrt{k} + d\eps^{2}}{2\left(1 - 2\sqrt{c_{d}}\right)k}\right) \label{eq:sum-NSNvecs-2}
    \\ = \ & \frac 14 \left(\frac{2\eps}{\sqrt{2\pi\left(1 + 2\sqrt{c_{d}} + 2c_{d}\right)k}}\right)^{d}\exp\left(-\frac{1 + \frac{2\eps}{\sqrt{k}} + \frac{d\eps^{2}}{k}}{2\left(1 - 2\sqrt{c_{d}}\right)}\right) \nonumber
    \\ \ge \ & \frac 14 \left(\frac{2\eps}{\sqrt{2\pi\left(1 + 2\sqrt{c_{d}} + 2c_{d}\right)k}}\right)^{d}\exp\left(-\frac{1 + \frac{1}{8} + \frac{1}{16}}{2\left(1 - 2\sqrt{c_{d}}\right)}\right) \label{eq:sum-NSNvecs-3}
    \\ \ge \ & \frac{1}{16} \left(\frac{2\eps}{\sqrt{2\pi\left(1 + 2\sqrt{c_{d}} + 2c_{d}\right)k}}\right)^{d}
    ,\label{eq:sum-NSNvecs-4}
  \end{align}
  where we have used that \(\normtwo*{\target} \le \normone*{\target} \le \sqrt{k}\) in \cref{eq:sum-NSNvecs-2}, that \(k \ge 16\), \(k \ge d\), and \(\eps < 1/4\) in \cref{eq:sum-NSNvecs-3}, and that
  \[
    \exp\left(-\frac{1 + \frac{1}{8} + \frac{1}{16}}{2\left(1 - 2\sqrt{c_{d}}\right)}\right) \ge \exp\left(-\frac{1 + \frac{1}{8} + \frac{1}{16}}{2\left(1 - 2\sqrt{\frac{1}{16}}\right)}\right) \ge \frac{1}{4}
  \]
  in \cref{eq:sum-NSNvecs-4}.
\end{proof}

Bounding $\expect{\numhits^2}$ requires handling stochastic dependencies.
Thus, we estimate the joint probability that two subsets of \(k\) elements of \(X_1, \ldots, X_n\) sum \(\eps\)-close to the same target, taking into account that the intersection of the subsets might not be empty.
The next lemma provides an upper bound on this joint probability that depends only on the size of the symmetric difference between the two subsets.

\begin{lemma}[Joint sum of NSN vectors]
  \label{lemma:sumNSN}
  Let $k, j \in \N$ with $j \le k$.
  Furthermore, let $X_{1}, \ldots, X_{k + j}$ be i.i.d.\ $d$-dimensional NSN
  random vectors with $k + j \ge Cd^{3}\log\frac{d}{\eps}$.
  Let
  $c_{d} = \min\left\{\frac{1}{d^{2}}, \frac{1}{16}\right\} $,
  $A = \sum_{i = 1}^{j}X_{i}$, $B = \sum_{i = j + 1}^{k}X_{i}$,
  and $C = \sum_{i = k + 1}^{k + j}X_{i}$.\footnote{We adopt the convention that $\sum_{i = 1}^{0}X_{i} = 0$.}
  Then, if \(j \ge 216\) it holds that
  \begin{align}
    & \pr{A + B \in \multiInterval{\eps}(\target), B + C \in \multiInterval{\eps}(\target)} \nonumber
    \\ \le \ & \frac{(2\eps)^{2d} \cdot \exp \left(\frac{2\eps^2d\left(k(1 - 2\sqrt{c_d}) - j(1 - 4\sqrt{c_d} - c_d)\right)}{j(1 - 2\sqrt{c_d})\left(2k(1 - 2\sqrt{c_d}) - j(1 - 6\sqrt{c_d} - 2 c_d)\right)}\right)}{(2\pi)^d(1 - 2\sqrt{c_d})^d \sqrt{\frac{j^d \left(2k(1 - 2\sqrt{c_d}) - j(1 - 6\sqrt{c_d} - 2 c_d)\right)^d}{(1 + 2\sqrt{c_d} + 2c_d)^d}}} + \exp[-(k-j)c_d] + 4\exp[-jc_d]
    .\label{eq:2ndmoment:1stcase}
  \end{align}
\end{lemma}
\begin{proof}
  Let \(n = k + j\).
  Since the $X_{i}$s are NSN random vectors, for each $i \in \left[n\right]$
  and $\ell \in \left[d\right]$ we can write the $\ell$-th entry of $X_{i}$
  as $(X_i)_l = Z_{i} \cdot Z_{i, \ell}$ where the variables in $\left\{Z_{i}\right\}_{i \in \left[n\right]}$
  and in $\left\{Z_{i, \ell}\right\}_{i \in \left[n\right], \ell \in \left[d\right]}$
  are i.i.d.\ random variables following a standard normal distribution.
  Let us first focus on \cref{eq:2ndmoment:1stcase}.
  Let $\cE$ be the intersection of the following events:
  \begin{enumerate}
    \item \(j\left(1 - 2\sqrt{c_{d}}\right) \le \sum_{i = 1}^{j}Z_{i}^{2} \le j \left(1 + 2\sqrt{c_d} + c_d\right)\).
    \item \(\sum_{i = j + 1}^{k}Z_{i}^{2} \ge (k-j)\left(1 - 2 \sqrt{c_d}\right)\).
    \item \(j\left(1 - 2\sqrt{c_{d}}\right) \le \sum_{i = k + 1}^{k + j}Z_{i}^{2} \le j \left(1 + 2\sqrt{c_d} + c_d\right)\)
  \end{enumerate}
  By the law of total probability, we have
  \begin{align}
    & \pr{A + B \in \multiInterval{\eps}(\target), B + C \in \multiInterval{\eps}(\target)} \nonumber
    \\ \le \ & \pr{A + B \in \multiInterval{\eps}(\target), B + C \in \multiInterval{\eps}(\target) \st \cE} + \pr{\overline{\cE}}\nonumber
    \\ = \ & \expectwrt{Z_{1}, \ldots, Z_{n}}{\pr{A + B \in \multiInterval{\eps}(\target), B + C \in \multiInterval{\eps}(\target) \st Z_{1}, \ldots, Z_{n}} \st \cE} + \pr{\overline{\cE}} \nonumber
    \\ = \ & \expectwrt{Z_{1}, \ldots, Z_{n}}{\prod_{i = 1}^{d}\pr{A_{i} + B_{i} \in \interval{\eps}{(\target_i)}, B_{i} + C_{i} \in \interval{\eps}{(\target_i)} \st Z_{1}, \ldots, Z_{n}}\st \cE} + \pr{\overline{\cE}} \label{eq:sumNSN-1}
    ,
  \end{align}
  where the latter equality holds by independence.

  Conditional on $Z_{1}, \ldots, Z_{n}$, we have that $A_{i} \sim \normal{0, \sum_{i = 1}^{j}Z_{i}^{2}}$, $B_{i} \sim \normal{0, \sum_{r = j + 1}^{k}Z_{r}^{2}}$, and $C_{i} \sim \normal{0, \sum_{i = k + 1}^{k + j}Z_{i}^{2}}$ for each $i \in [d]$.
  Furthermore, conditional on $Z_{1}, \ldots, Z_{n}$, it holds that \(A_i, B_i, C_i\) are centred, independent random variables.
  Hence, \cref{lemma:tool:integral1-MRSSP:clever} (\cref{app:tools}) applies and implies that
  \begin{align*}
    & \pr{A_{i} + B_{i} \in \interval{\eps}{(\target_i)}, B_{i} + C_{i} \in \interval{\eps}{(\target_i)} \st Z_{1}, \ldots, Z_{n}}
    \\ \le \ & \pr{A_{i} + B_{i} \in \interval{\eps}{(0)}, B_{i} + C_{i} \in \interval{\eps}{(0)} \st Z_{1}, \ldots, Z_{n}}
  \end{align*}
  Then,
  \begin{align}
    & \pr{A_{i} + B_{i} \in \interval{\eps}{(0)}, B_{i} + C_{i} \in \interval{\eps}{(0)} \st Z_{1}, \ldots, Z_{n}} \nonumber
    \\ = \ & {\pr{A_{i} \in \interval{\eps}{(-B_{i})}, C_{i} \in \interval{\eps}{(-B_{i})} \st Z_{1}, \ldots, Z_{n}}} \nonumber
    \\ = \ & \int_{\R} \varphi_{B_i}(x) \left(\int_{-\eps}^{+\eps} \varphi_{A_i} (x + s) \dd{s}\right)\left(\int_{-\eps}^{+\eps} \varphi_{C_i} (x + s) \dd{s}\right) \dd{x}, \nonumber
  \end{align}
  where \(\varphi_{A_i}, \varphi_{B_i}, \varphi_{C_i}\) are the density functions of, respectively, \(A_i, B_i, C_i\) conditional on \(Z_{1}, \ldots, Z_{n}\).
  Since \(2 \sum_{i = 1}^{j} Z_i^2 \ge j(1 - 2c_d) \ge 162\) and \(2 \sum_{i = k + 1}^{k + j} Z_i^2 \ge j(1 - 2c_d) \ge 162\), \cref{lemma:tool:integral2-MRSSP} (\cref{app:tools}) holds, and we have
  \begin{align*}
    & \left(\int_{-\eps}^{+\eps} \varphi_{A_i}(x + s) \dd{s}\right)\left(\int_{-\eps}^{+\eps} \varphi_{C_i}(x + s) \dd{s}\right)
    \\ \le \ & \left[\int_{-\eps}^{+\eps} \frac{\exp\left(-\frac{(x + \eps)^2}{2 \sum_{i = 1}^{j} Z_i^2}\right) + \exp\left(-\frac{(x - \eps)^2}{2\sum_{i = 1}^{j} Z_i^2}\right)}{2 \sqrt{2 \pi \sum_{i = 1}^{j} Z_i^2}} \cdot \exp\left(\frac{\eps^2}{2\sum_{i = 1}^{j} Z_i^2}\right) \dd{s}\right]
    \\ & \cdot \left[\int_{-\eps}^{+\eps} \frac{\exp\left(-\frac{(x + \eps)^2}{2 \sum_{i = k + 1}^{k + j} Z_i^2}\right) + \exp\left(-\frac{(x - \eps)^2}{2\sum_{i = k + 1}^{k + j} Z_i^2}\right)}{2 \sqrt{2 \pi \sum_{i = k + 1}^{k + j} Z_i^2}} \cdot \exp\left(\frac{\eps^2}{2\sum_{i = k + 1}^{k + j} Z_i^2}\right) \dd{s}\right]
    .
  \end{align*}
  Since the expectation we wish to compute (\cref{eq:sumNSN-1}) is conditional on \(\cE\), we can exploit the fact that \(j(1 - 2\sqrt{c_d}) \le \sum_{i = 1}^j Z_i^2 \le j(1 + 2\sqrt{c_d} + 2c_d)\) and \(j(1 - 2\sqrt{c_d}) \le \sum_{i = k + 1}^{k + j} Z_i^2 \le j(1 + 2\sqrt{c_d} + 2c_d)\), obtaining the following:
  \begin{align*}
    & \left[\int_{-\eps}^{+\eps} \frac{\exp\left(-\frac{(x + \eps)^2}{2 \sum_{i = 1}^{j} Z_i^2}\right) + \exp\left(-\frac{(x - \eps)^2}{2\sum_{i = 1}^{j} Z_i^2}\right)}{2 \sqrt{2 \pi \sum_{i = 1}^{j} Z_i^2}} \cdot \exp\left(\frac{\eps^2}{2\sum_{i = 1}^{j} Z_i^2}\right) \dd{s}\right]
    \\ & \cdot \left[\int_{-\eps}^{+\eps} \frac{\exp\left(-\frac{(x + \eps)^2}{2 \sum_{i = k + 1}^{k + j} Z_i^2}\right) + \exp\left(-\frac{(x - \eps)^2}{2\sum_{i = k + 1}^{k + j} Z_i^2}\right)}{2 \sqrt{2 \pi \sum_{i = k + 1}^{k + j} Z_i^2}} \cdot \exp\left(\frac{\eps^2}{2\sum_{i = k + 1}^{k + j} Z_i^2}\right) \dd{s}\right]
    \\ \le \ & \left[\int_{-\eps}^{+\eps} \frac{\exp\left(-\frac{(x + \eps)^2}{2j(1 + 2\sqrt{c_d} + 2c_d)}\right) + \exp\left(-\frac{(x - \eps)^2}{2j(1 + 2\sqrt{c_d} + 2c_d)}\right)}{2 \sqrt{2 \pi j(1 - 2\sqrt{c_d})}} \cdot \exp\left(\frac{\eps^2}{2j(1 - 2\sqrt{c_d})}\right) \dd{s}\right]
    \\ & \cdot \left[\int_{-\eps}^{+\eps} \frac{\exp\left(-\frac{(x + \eps)^2}{2 j(1 + 2\sqrt{c_d} + 2c_d)}\right) + \exp\left(-\frac{(x - \eps)^2}{2j(1 + 2\sqrt{c_d} + 2c_d)}\right)}{2 \sqrt{2 \pi j(1 - 2\sqrt{c_d})}} \cdot \exp\left(\frac{\eps^2}{2j(1 - 2\sqrt{c_d})}\right) \dd{s}\right]
    \\ = \ & \left[\int_{-\eps}^{+\eps} \frac{\exp\left(-\frac{(x + \eps)^2}{2j(1 + 2\sqrt{c_d} + 2c_d)}\right) + \exp\left(-\frac{(x - \eps)^2}{2j(1 + 2\sqrt{c_d} + 2c_d)}\right)}{2 \sqrt{2 \pi j(1 - 2\sqrt{c_d})}} \cdot \exp\left(\frac{\eps^2}{2j(1 - 2\sqrt{c_d})}\right) \dd{s}\right]^2
    .
  \end{align*}
  Then,
  \begin{align*}
    & \left[\int_{-\eps}^{+\eps} \frac{\exp\left(-\frac{(x + \eps)^2}{2j(1 + 2\sqrt{c_d} + 2c_d)}\right) + \exp\left(-\frac{(x - \eps)^2}{2j(1 + 2\sqrt{c_d} + 2c_d)}\right)}{2 \sqrt{2 \pi j(1 - 2\sqrt{c_d})}} \cdot \exp\left(\frac{\eps^2}{2j(1 - 2\sqrt{c_d})}\right) \dd{s}\right]^2
    \\ = \ & \frac{(2\eps)^2 \exp\left(\frac{\eps^2}{j(1 - 2\sqrt{c_d})}\right)}{8 \pi j(1 - 2\sqrt{c_d})} \cdot \Bigg[\exp\left(-\frac{(x + \eps)^2}{j(1 + 2\sqrt{c_d} + 2c_d)}\right) + 2\exp \left(-\frac{x^2 + \eps^2}{j(1 + 2\sqrt{c_d} + 2c_d)}\right)
    \\ & + \exp \left(-\frac{(x-\eps)^2}{j(1 + 2\sqrt{c_d} + 2c_d)}\right) \Bigg]
    .
  \end{align*}
  Let \(\varphi_Y\) be the density function of a random variable \(Y \sim \normal{0, j(1 + 2\sqrt{c_d} + 2c_d)}\).
  Then,
  \begin{align*}
    & \frac{(2\eps)^2 \exp\left(\frac{\eps^2}{j(1 - 2\sqrt{c_d})}\right)}{8 \pi j(1 - 2\sqrt{c_d})} \cdot \Bigg[\exp\left(-\frac{(x + \eps)^2}{j(1 + 2\sqrt{c_d} + 2c_d)}\right) + 2\exp \left(-\frac{x^2 + \eps^2}{j(1 + 2\sqrt{c_d} + 2c_d)}\right)
    \\ & + \exp \left(-\frac{(x-\eps)^2}{j(1 + 2\sqrt{c_d} + 2c_d)}\right) \Bigg]
    \\ = \ & \frac{\eps^2 \sqrt{1 + 2\sqrt{c_d} + 2c_d} \cdot \exp \left(\frac{\eps^2}{j(1 - 2\sqrt{c_d})}\right)}{2\sqrt{\pi j}(1 - 2\sqrt{c_d})}
    \\ & \cdot \left(\varphi_{\frac{Y}{\sqrt{2}}}(x + \eps) + \varphi_{\frac{Y}{\sqrt{2}}}(x - \eps) + 2\exp\left(-\frac{\eps^2}{j(1 + 2\sqrt{c_d} + 2c_d)}\right) \varphi_{\frac{Y}{\sqrt{2}}}(x)\right)
    .
  \end{align*}
  Furthermore, it holds that
  \begin{align*}
    & \int_{\R} \varphi_{B_i} (x) \left(\varphi_{\frac{Y}{\sqrt{2}}}(x + \eps) + \varphi_{\frac{Y}{\sqrt{2}}}(x - \eps) + 2\exp\left(-\frac{\eps^2}{j(1 + 2\sqrt{c_d} + 2c_d)}\right) \varphi_{\frac{Y}{\sqrt{2}}}(x)\right) \dd{x}
    \\ = \ & (\varphi_{B_i} \ast \varphi_{\frac{Y}{\sqrt{2}}})(+\eps) + (\varphi_{B_i} \ast \varphi_{\frac{Y}{\sqrt{2}}})(-\eps) + 2\exp\left(-\frac{\eps^2}{j(1 + 2\sqrt{c_d} + 2c_d)}\right) (\varphi_{B_i} \ast \varphi_{\frac{Y}{\sqrt{2}}})(0)
    \\ = \ & (\varphi_{B_i + \frac{Y}{\sqrt{2}}})(+\eps) + (\varphi_{B_i + \frac{Y}{\sqrt{2}}})(-\eps) + 2\exp\left(-\frac{\eps^2}{j(1 + 2\sqrt{c_d} + 2c_d)}\right) (\varphi_{B_i + \frac{Y}{\sqrt{2}}})(0)
    ,
  \end{align*}
  where \(\ast\) denotes the convolution operation, while \(\varphi_{B_i + \frac{Y}{\sqrt{2}}}\) denote the density function of \(B_i + \frac{Y}{\sqrt{2}}\) (conditional on \(Z_1, \ldots, Z_n\)).

  Then,
  \begin{align*}
    & (\varphi_{B_i + \frac{Y}{\sqrt{2}}})(+\eps) + (\varphi_{B_i + \frac{Y}{\sqrt{2}}})(-\eps) + 2\exp\left(-\frac{\eps^2}{j(1 + 2\sqrt{c_d} + 2c_d)}\right) (\varphi_{B_i + \frac{Y}{\sqrt{2}}})(0)
    \\ \le \ & \frac{2\exp \left(-\frac{\eps^2}{j(1 + 2\sqrt{c_d} + 2c_d) + 2\sum_{i = j + 1}^k Z_i^2}\right) + 2\exp\left(-\frac{\eps^2}{j(1 + 2\sqrt{c_d} + 2c_d)}\right)}{\sqrt{\pi \left(j(1 + 2\sqrt{c_d} + 2c_d) + 2\sum_{i = j + 1}^k Z_i^2\right)}}
    \\ \le \ & \frac{4\exp \left(-\frac{\eps^2}{j(1 + 2\sqrt{c_d} + 2c_d) + 2\sum_{i = j + 1}^k Z_i^2}\right)}{\sqrt{\pi \left(j(1 + 2\sqrt{c_d} + 2c_d) + 2\sum_{i = j + 1}^k Z_i^2\right)}}
    .
  \end{align*}

  Hence,
  \begin{align}
    & \pr{A_{i} \in \interval{\eps}{(-B_{i})}, C_{i} \in \interval{\eps}{(-B_{i})} \st Z_{1}, \ldots, Z_{n}} \nonumber
    \\ \le \ & \frac{\eps^2 \sqrt{1 + 2\sqrt{c_d} + 2c_d} \cdot \exp \left(\frac{\eps^2}{j(1 - 2\sqrt{c_d})}\right)}{\sqrt{\pi j}(1 - 2\sqrt{c_d})} \cdot \frac{2\exp \left(-\frac{\eps^2}{j(1 + 2\sqrt{c_d} + 2c_d) + 2\sum_{i = j + 1}^k Z_i^2}\right)}{\sqrt{\pi \left(j(1 + 2\sqrt{c_d} + 2c_d) + 2\sum_{i = j + 1}^k Z_i^2\right)}}.\label{eq:2ndmoment:convfinal}
  \end{align}
  Again, since we wish to compute an expectation that is conditional on \(\cE\) (\cref{eq:sumNSN-1}), we exploit the fact that the maximum of \cref{eq:2ndmoment:convfinal} is reached when \(\sum_{i = j + 1}^{k}Z_{i}^{2} = (k-j)\left(1 - 2 \sqrt{c_d}\right) \) (see \cref{lemma:find-maximum:2} in \cref{app:tools}).
  Hence,
  \begin{align}
    & \pr{A_{i} \in \interval{\eps}{(-B_{i})}, C_{i} \in \interval{\eps}{(-B_{i})} \st Z_{1}, \ldots, Z_{n}} \nonumber
    \\ \le \ & \frac{2\eps^2 \sqrt{1 + 2\sqrt{c_d} + 2c_d} \cdot \exp \left(\frac{\eps^2}{j(1 - 2\sqrt{c_d})}\right)}{\sqrt{\pi j}(1 - 2\sqrt{c_d})} \cdot \frac{\exp \left(-\frac{\eps^2}{j(1 + 2\sqrt{c_d} + 2c_d) + 2(k-j)(1 - 2\sqrt{c_d})}\right)}{\sqrt{\pi \left(j(1 + 2\sqrt{c_d} + 2c_d) + 2(k-j)(1 - 2\sqrt{c_d})\right)}} \nonumber
    \\ = \ & \frac{2\eps^2 \sqrt{1 + 2\sqrt{c_d} + 2c_d} \cdot \exp \left(\frac{2\eps^2\left(k(1 - 2\sqrt{c_d}) - j(1 - 4\sqrt{c_d} - c_d)\right)}{j(1 - 2\sqrt{c_d})\left(2k(1 - 2\sqrt{c_d}) - j(1 - 6\sqrt{c_d} - 2 c_d)\right)}\right)}{\pi(1 - 2\sqrt{c_d})\sqrt{j \left(2k(1 - 2\sqrt{c_d}) - j(1 - 6\sqrt{c_d} - 2 c_d)\right)}}
    .\label{eq:2ndmoment:convfinal2}
  \end{align}
  By the law of total probability,
  \begin{align*}
    & \pr{A + B \in \multiInterval{\eps}(\target), B + C \in \multiInterval{\eps}(\target)}
    \\ \le \ & \pr{A + B \in \multiInterval{\eps}(\target), B + C \in \multiInterval{\eps}(\target) \st \cE} + \pr{\overline{\cE}}
    \\ \le \ & \expectwrt{Z_{1}, \ldots, Z_{n}}{\prod_{i = 1}^{d}\pr{A_{i} + B_{i}, B_{i} + C_{i} \in \interval{\eps}{(\target_i)} \st Z_{1}, \ldots, Z_{n}} \st \cE} + \pr{\overline{\cE}}
    \\ \le \ & \frac{(2\eps)^{2d} \sqrt{(1 + 2\sqrt{c_d} + 2c_d)^d} \cdot \exp \left(\frac{2\eps^2d\left(k(1 - 2\sqrt{c_d}) - j(1 - 4\sqrt{c_d} - c_d)\right)}{j(1 - 2\sqrt{c_d})\left(2k(1 - 2\sqrt{c_d}) - j(1 - 6\sqrt{c_d} - 2 c_d)\right)}\right)}{(2\pi)^d(1 - 2\sqrt{c_d})^d \sqrt{j^d \left(2k(1 - 2\sqrt{c_d}) - j(1 - 6\sqrt{c_d} - 2 c_d)\right)^d}} + \pr{\overline{\cE}}
    .
  \end{align*}
  By \cref{lemma:chisquared_bound} and the union bound, the probability that \(\cE\) occurs is at least \(1 - \exp[-(k-j)c_d] - 4\exp[-jc_d]\).
  Thus,
  \begin{align*}
    & \pr{A + B \in \multiInterval{\eps}(\target), B + C \in \multiInterval{\eps}(\target)}
    \\ \le \ & \frac{(2\eps)^{2d} \sqrt{(1 + 2\sqrt{c_d} + 2c_d)^d} \cdot \exp \left(\frac{2\eps^2d\left(k(1 - 2\sqrt{c_d}) - j(1 - 4\sqrt{c_d} - c_d)\right)}{j(1 - 2\sqrt{c_d})\left(2k(1 - 2\sqrt{c_d}) - j(1 - 6\sqrt{c_d} - 2 c_d)\right)}\right)}{(2\pi)^d(1 - 2\sqrt{c_d})^d \sqrt{j^d \left(2k(1 - 2\sqrt{c_d}) - j(1 - 6\sqrt{c_d} - 2 c_d)\right)^d}}
    \\ + \ & \exp[-(k-j)c_d] + 4\exp[-jc_d]
    .
  \end{align*}

\end{proof}

The following lemma provides an explicit expression for the variance of the $\eps$-subset-sum number.

\begin{lemma}[Second moment of $\numhits$]
  \label{lemma:secondmomnet_subsetsumnumber}Let $k, n$ be positive integers.
  Let \(S_0, S_1, \ldots, S_k\) be subsets of \([n]\) such that \(\abs{S_0 \cap S_j} = k-j\) for \(j = 0, 1, \ldots, k\).
  Let \(\randSetOne, \randSetTwo\) be two random variables yielding two subsets of \([n]\) of size \(k\) drawn independently and uniformly at random.
  Let $X_{1}, \ldots, X_{n}$ be $d$-dimensional i.i.d.\ NSN random vectors.
  For any $\eps > 0$ and $\target \in \R^{d}$, the second moment
  of the $(\eps, k)$-subset-sum number is
  \[
    \expect{\numhits^{2}} = \binom{n}{k}^{2} \sum_{j = 0}^{k} \pr{\cH_{S_0}\cap\cH_{S_j}} \pr{\abs{\randSetOne \cap \randSetTwo} = k - j}
    ,
  \]
  where $\cH_S$ denotes the event $\normmax*{\left(\sum_{i \in S}X_{i}\right)-\target} \le \eps$.
\end{lemma}

\begin{proof}
  Let \(\randSetOne, \randSetTwo\) two random variables yielding two elements of \(\binom{[n]}{k}\) drawn independently and uniformly at random.
  By the definition of \(\numhits\), we have that
  \begin{align*}
    \expect{\numhits^{2}}
     & = \expect{\biggl(\sum_{S \in \binom{[n]}{k}} \ind_{\cH_S}\biggr)\biggl(\sum_{T \in \binom{[n]}{k}}\ind_{\cH_T}\biggr)}
    \\ & = \expect{\sum_{S, T \in \binom{[n]}{k}} \ind_{\cH_S}\ind_{\cH_T}}
    \\ & = \sum_{S, T \in \binom{[n]}{k}} \pr{\cH_S \cap \cH_T}
    \\ & = \sum_{S, T \in \binom{[n]}{k}} \pr{\cH_{\randSetOne} \cap \cH_{\randSetTwo} \st \randSetOne = S, \randSetTwo = T } \pr{\randSetOne = S, \randSetTwo = T}
    \\ & = \binom{n}{k}^{2} \sum_{j = 0}^k \pr{\cH_{\randSetOne}\cap\cH_{\randSetTwo} \st \abs{\randSetOne \cap \randSetTwo} = k - j} \pr{\abs{\randSetOne \cap \randSetTwo} = k - j}
    ,
  \end{align*}
  as \(\pr{\cH_{\randSetOne}\cap\cH_{\randSetTwo}}\) depends only on the size of \(\randSetOne \cap \randSetTwo\).
  The thesis follows by observing that
  \[
    \pr{\cH_{\randSetOne}\cap\cH_{\randSetTwo} \st \abs{\randSetOne \cap \randSetTwo} = k - j} = \pr{\cH_{S_0}\cap\cH_{S_j}}
    .
  \]
\end{proof}

\subsubsection*{Proof of \cref{thm:mrrs_gaussian_scaled}}
  We use the second moment method (\cref{lemma:secondmoment} in \cref{app:tools}) on
  the {$\eps$-subset-sum number} $\numhits$ of $X_{1}, \ldots, X_{n}$.
  Thus, we aim to provide a lower bound on the right-hand side of
  \begin{align}
    \pr{T > 0}
    \ge \frac{\expect{\numhits}^2}{\expect{\numhits^2}}
    .\label{eq:secondmoment}
  \end{align}
  Equivalently, we can provide an upper bound on the inverse $\frac{\expect{\numhits^{2}}}{\expect{\numhits}^{2}}$.
  By \cref{lemma:secondmomnet_subsetsumnumber}
  \begin{align}
    \expect{\numhits^{2}} = \binom{n}{k}^{2}\sum_{j = 0}^{k}\pr{\abs{\randSetOne\cap\randSetTwo} = k - j}\pr{\cH_{S_{0}}\cap\cH_{S_{j}}}\label{eq:secondmoment_decomp}
  \end{align}
  where $\randSetOne, \randSetTwo, S_{i}$ and $\cH$ are defined as in the statement
  of the lemma.
  Observe also that
  \begin{align}
    \expect{\numhits}
    = \sum_{S \in \binom{[n]}{k}} \expect{\ind_{\cH}}
    = \sum_{S \in \binom{[n]}{k}} \pr{\cH}
    = \binom{n}{k} \pr{\cH_{S_{0}}}
    .\label{eq:expected_hits}
  \end{align}
  By \cref{eq:secondmoment_decomp,eq:expected_hits},
  \begin{align}
    \frac{\expect{\numhits^{2}}}{\expect{\numhits}^{2}}
     & = \frac{\binom{n}{k}^{2}}{\expect{\numhits}^{2}}\sum_{j = 0}^{k}\pr{\abs{\randSetOne\cap\randSetTwo} = k - j}\pr{\cH_{S_{0}}\cap\cH_{S_{j}}}\nonumber
    \\ & = \sum_{j = 0}^{k}\pr{\abs{\randSetOne\cap\randSetTwo} = k - j}\frac{\pr{\cH_{S_{0}}\cap\cH_{S_{j}}}}{\pr{\cH_{S_{0}}}^{2}} \nonumber
    .
  \end{align}
  Let \(k = \alpha n\), and let \(j^\star = \Ceil*{k-2\alpha^2 n} = \Ceil*{\alpha n(1 - 2\alpha)}\), which implies \(k - j^\star \le 2\alpha^2 n\).
  The term \(\frac{\pr{\cH_{S_0} \cap \cH_{S_{j}}}}{\pr{\cH_{S_0}}^2}\) is non-decreasing in the intersection size (\cref{lemma:monotonicity:intersection} in \cref{app:tools}).
  Hence, we can split the sum in two parts as follows:
  \begin{align}
    & \sum_{j = 0}^{k}\pr{\abs{\randSetOne\cap\randSetTwo} = k - j}\frac{\pr{\cH_{S_{0}}\cap\cH_{S_{j}}}}{\pr{\cH_{S_{0}}}^{2}} \nonumber
    \\ = \ & \sum_{j = 0}^{j^\star - 1}\pr{\abs{\randSetOne\cap\randSetTwo} = k - j}\frac{\pr{\cH_{S_{0}}\cap\cH_{S_{j}}}}{\pr{\cH_{S_{0}}}^{2}} + \sum_{j = j^\star}^{k}\pr{\abs{\randSetOne\cap\randSetTwo} = k - j}\frac{\pr{\cH_{S_{0}}\cap\cH_{S_{j}}}}{\pr{\cH_{S_{0}}}^{2}} \nonumber
    \\ \le \ & \pr{\abs{\randSetOne \cap \randSetTwo} \ge 2\alpha^2 n} \cdot \frac{1}{\pr{\cH_{S_0}}} + \frac{\pr{\cH_{S_0} \cap \cH_{S_{j^\star}}}}{\pr{\cH_{S_0}}^2}
    .\label{eq:secondmoment_bound}
  \end{align}
  As for the denominator in \cref{eq:secondmoment_bound},
  by \cref{lemma:prob_one_set} we have
  \begin{align}
    \pr{\cH_{S_{0}}}^{2} \ge \frac{1}{256}\left(\frac{4\eps^{2}}{2\pi\left(1 + 2\sqrt{c_{d}} + 2c_{d}\right)k}\right)^{d}
    .\label{eq:denominator_bound}
  \end{align}
  As for the numerator of the first term in \cref{eq:secondmoment_bound}, \cref{lemma:sumNSN} gives us two different bounds.
  Now, let \(\alpha = k/n\).
  In order to estimate
  \[
    \frac{\pr{\cH_{S_{0}}\cap\cH_{S_{j}}}}{\pr{\cH_{S_{0}}}^{2}}
  \]
  we first assume \(k - j \le 2\alpha^2 n\).
  As \(\pr{\cH_{S_{0}}\cap\cH_{S_{j}}}\) increases with the intersection size (\cref{lemma:monotonicity:intersection}), we bound the term from above by \({\pr{\cH_{S_{0}}\cap\cH_{S_{j^\star}}}}\) with \(j^\star = \Ceil*{k - 2\alpha^2 n} = \Ceil*{\alpha n (1 - 2\alpha)}\).
  In this case, the intersection size is \(\Floor*{2\alpha^2 n}\).
  Then, by \cref{eq:2ndmoment:1stcase},
  \begin{align*}
    & \pr{\cH_{S_{0}}\cap\cH_{S_{j^\star}}}
    \\ \le \ & \frac{(2\eps)^{2d} \cdot \exp \left(\frac{2\eps^2d\left(\alpha n (1 - 2\sqrt{c_d}) - \alpha n (1 - 2\alpha)(1 - 4\sqrt{c_d} - c_d)\right)}{\alpha n (1 - 2\alpha)(1 - 2\sqrt{c_d})\left(2\alpha n(1 - 2\sqrt{c_d}) - \Ceil*{\alpha n (1 - 2\alpha)}(1 - 6\sqrt{c_d} - 2 c_d)\right)}\right)}{(2\pi)^d(1 - 2\sqrt{c_d})^d \sqrt{(\alpha n (1 - 2\alpha))^d \left(\frac{2\alpha n(1 - 2\sqrt{c_d}) - \Ceil*{\alpha n (1 - 2\alpha)}(1 - 6\sqrt{c_d} - 2 c_d)}{1 + 2\sqrt{c_d} + 2c_d}\right)^d}}
    \\ &+ \exp[-(\alpha n-\Ceil*{\alpha n (1 - 2\alpha)})c_d] + 4\exp[-\alpha n (1 - 2\alpha)c_d]
    \\ \le \ & \frac{(2\eps)^{2d} \cdot \exp \left(\frac{2\eps^2d\left(2\sqrt{c_d} + \alpha + c_d - 8\alpha\sqrt{c_d} -2\alpha c_d\right)}{\alpha n (1 - 2\alpha)(1 - 2\sqrt{c_d})\left((1 + 2\sqrt{c_d} + 2c_d + 2\alpha -12\alpha \sqrt{c_d} -4\alpha c_d) - \frac{1 - 6\sqrt{c_d} - 2c_d}{\alpha n}\right)}\right)}{(2\alpha n \pi)^d(1 - 2\sqrt{c_d})^d \sqrt{(1 - 2\alpha)^d \left(\frac{(1 + 2\sqrt{c_d} + 2c_d + 2\alpha -12\alpha \sqrt{c_d} -4\alpha c_d) - \frac{1 - 6\sqrt{c_d} - 2c_d}{\alpha n}}{1 + 2\sqrt{c_d} + 2c_d}\right)^d}}
    \\ &+ \exp[(1 - 2 \alpha^2 n)c_d] + 4\exp[-\alpha n (1 - 2\alpha)c_d]
  \end{align*}
  Hence,
  \begin{align*}
    & \frac{\pr{\cH_{S_{0}}\cap\cH_{S_{j^\star}}}}{\pr{\cH_{S_0}}^2}
    \\ \le \ & \frac{256(1 + 2\sqrt{c_d} + 2c_d)^d \cdot \exp \left(\frac{2\eps^2d\left(2\sqrt{c_d} + \alpha + c_d - 8\alpha\sqrt{c_d} -2\alpha c_d\right)}{\alpha n (1 - 2\alpha)(1 - 2\sqrt{c_d})\left((1 + 2\sqrt{c_d} + 2c_d + 2\alpha -12\alpha \sqrt{c_d} -4\alpha c_d) - \frac{1 - 6\sqrt{c_d} - 2c_d}{\alpha n}\right)}\right)}{(1 - 2\sqrt{c_d})^d \sqrt{(1 - 2\alpha)^d \left(\frac{(1 + 2\sqrt{c_d} + 2c_d + 2\alpha -12\alpha \sqrt{c_d} -4\alpha c_d) - \frac{1 - 6\sqrt{c_d} - 2c_d}{\alpha n}}{1 + 2\sqrt{c_d} + 2c_d}\right)^d}}
    \\ &+ 256\exp\left[d \log \frac{2 \pi \alpha n}{4 \eps^2} + d\log (1 + 2\sqrt{c_d} + 2c_d)\right]\left(\exp[(1 - 2 \alpha^2 n)c_d] + 4\exp[-\alpha n (1 - 2\alpha)c_d]\right).
  \end{align*}
  Let us fix \(\alpha = 1/(6\sqrt{d})\), and consider the term
  \begin{align*}
    & (1 - 2\alpha)^d \left(\frac{(1 + 2\sqrt{c_d} + 2c_d + 2\alpha -12\alpha \sqrt{c_d} -4\alpha c_d) - \frac{1 - 6\sqrt{c_d} - 2c_d}{\alpha n}}{1 + 2\sqrt{c_d} + 2c_d}\right)^d
    \\ = \ & \left[\frac{1 - 4\alpha^2 + (1 - 2\alpha)\left(2\sqrt{c_d}(1 - 6\alpha) + 2c_d(1 - 2\alpha) - \frac{1 - 6\sqrt{c_d} - 2c_d}{\alpha n}\right)}{1 + 2\sqrt{c_d} + 2c_d}\right]^d
    .
  \end{align*}
  Observe that \(1 - 6\alpha \ge 0\), \(1 - 2\alpha \ge \frac{2}{3}\), \(1 - 6\sqrt{c_d} - 2c_d \le 1\), and \(1 + 2\sqrt{c_d} + 2c_d \le 1 + \frac{4}{d}\).
  Assume \(n \ge d^{3/2} \log \frac 1\eps\).
  Hence,
  \begin{align}
    & \left[\frac{1 - 4\alpha^2 + (1 - 2\alpha)\left(2\sqrt{c_d}(1 - 6\alpha) + 2c_d(1 - 2\alpha) - \frac{1 - 6\sqrt{c_d} - 2c_d}{\alpha n}\right)}{1 + 2\sqrt{c_d} + 2c_d}\right]^d \nonumber
    \\ \ge \ & \left[\frac{1 - \frac{1}{9d} - \frac{2}{3d \log \frac 1\eps}}{1 + \frac{4}{d}}\right]^d \nonumber
    \\ \ge \ & \left[\frac{1 - \frac{7}{9d}}{1 + \frac{4}{d}}\right]^d \nonumber
    \\ = \ & \left(1 - \frac{43}{9(d + 4)}\right)^{d} \nonumber
    \\ \ge \ & a_1
    ,\label{eq:thmRSSP:term1}
  \end{align}
  where \(a_1\) is a positive constant.
  To justify the latter inequality, let
  \[
    f(d) = \left(1 - \frac{43}{9(d + 4)}\right)^{d}
    .
  \]
  Observe that \(f(d) > 0\) for all \(d \ge 1\).
  Furthermore, it is well known that \(\lim_{d t_1 \infty} f(d) = \exp(\frac{43}{9})\) and that \(f(d)\) is continuous in \([1, + \infty)\).
  Thus, there exists a constant \(a_1 > 0\) such that \(f(d) \ge a_1\) for all \(d \ge 1\).
  Now we take care of the term
  \begin{align*}
    \left(\frac{1 + 2\sqrt{c_d} + 2c_d}{1 - 2\sqrt{c_d}}\right)^d
    .
  \end{align*}
  Notice that
  \begin{align}
    \left(\frac{1 + 2\sqrt{c_d} + 2c_d}{1 - 2\sqrt{c_d}}\right)^d \nonumber
     & = \left(1 + \frac{4\sqrt{c_d} + 2c_d}{1 - 2\sqrt{c_d}}\right)^d \nonumber
    \\ & \le \left(1 + \frac{\frac{4}{d} + \frac{2}{d^2}}{1 - \frac{2}{d}}\right)^d \nonumber
    \\ & = \left(1 + \frac{4 + \frac{2}{d}}{d - 2}\right)^d \nonumber
    \\ & \le \left(1 + \frac{6}{d - 2}\right)^d \nonumber
    \\ & \le a_2
    ,\label{eq:thmRSSP:term2}
  \end{align}
  where the latter inequality holds with \(a_2\) being a positive constant for reasons that are analogous to those in \cref{eq:thmRSSP:term1}.
  Consider now the first exponential term.
  We have the following:
  \begin{align*}
    & \exp \left[\frac{2\eps^2d\left(2\sqrt{c_d} + \alpha + c_d - 8\alpha\sqrt{c_d} -2\alpha c_d\right)}{\alpha n (1 - 2\alpha)(1 - 2\sqrt{c_d})\left((1 + 2\sqrt{c_d} + 2c_d + 2\alpha -12\alpha \sqrt{c_d} -4\alpha c_d) - \frac{1 - 6\sqrt{c_d} - 2c_d}{\alpha n}\right)}\right]
    \\ \le \ & \exp \left[\frac{2\eps^2d}{\alpha n \cdot \frac{2}{3} \cdot \frac{1}{2} \cdot \left(1 - \frac{1}{\alpha n}\right)}\right]
    ,
  \end{align*}
  since \(2\sqrt{c_d} - 12\alpha \sqrt{c_d} \ge 0\) and \(2\alpha - 4\alpha c_d \ge 0\).
  Furthermore, as \(\alpha n \ge 2\), we get
  \begin{align*}
    \exp \left[\frac{2\eps^2d}{\frac 13 \alpha n \cdot \left(1 - \frac{1}{\alpha n}\right)}\right]
    \le \exp \left[\frac{12\eps^2d}{\alpha n}\right]
    .
  \end{align*}
  Now, if \(\alpha n \ge 12\eps^2 d\), we can bound the entire term from above by \(e\).
  Hence, \(n \ge 72 d^{\frac 32} \) is sufficient to get
  \begin{align}
    \exp \left[\frac{12\eps^2d}{\alpha n}\right] \le e
    .\label{eq:thmRSSP:term3}
  \end{align}
  Consider \(\exp\left[d \log \frac{\pi \alpha n}{4\eps^2} + d\log(1 + 2\sqrt{c_d} + 2c_d) + (1 - 2\alpha^2 n) c_d\right]\).
  It holds that
  \begin{align*}
    & \exp\left[d \log \frac{2\pi \alpha n}{4\eps^2} + d\log(1 + 2\sqrt{c_d} + 2c_d) + (1 - 2\alpha^2 n) c_d\right]
    \\ \le \ & \exp\left[d \log \frac{13\pi \alpha n}{16\eps^2} - \frac{\alpha^2 n}{8d^2} + 1\right]
    \\ \le \ & \exp\left[d \log \frac{13\pi n}{96 \sqrt{d} \eps^2} - \frac{n}{288d^3} + 1\right]
    .
  \end{align*}
  If \(n \ge a_3d^4(\log d + \log \frac{1}{\eps})\) for a large enough constant \(a_3 > 0\), we have that
  \begin{align}
    \exp\left[d \log \frac{13\pi n}{96 \sqrt{d} \eps^2} - \frac{n}{288d^3} + 1\right] < \frac{1}{256}
    .\label{eq:thmRSSP:term4}
  \end{align}
  Analogously, if \(n \ge a_4 d^{3.5}(\log d + \log \frac{1}{\eps}) \) for a large enough constant \(a_4 > 0\), we have that
  \begin{align}
    \exp\left[d \log \frac{2\pi \alpha n}{4\eps^2} + d\log(1 + 2\sqrt{c_d} + 2c_d) - {\alpha n (1 - 2\alpha)c_d}\right] \le \frac{1}{4 \cdot 256}
    .\label{eq:thmRSSP:term5}
  \end{align}
  Altogether, \cref{eq:thmRSSP:term1,eq:thmRSSP:term2,eq:thmRSSP:term3,eq:thmRSSP:term4} imply that, for all \(j\) such that \(k - j \le 2\alpha^2 n\),
  \begin{align}
    \frac{\pr{\cH_{S_{0}}\cap\cH_{S_{j}}}}{\pr{\cH_{S_0}}^2} \le 256 \cdot \frac{a_2}{a_1} \cdot e + 2
    .\label{eq:mrssp:small-intersection}
  \end{align}

  Now we bound the probability \(\pr{\abs{\randSetOne\cap\randSetTwo} \ge 2\alpha^2 n}\).
  Notice that
  \begin{align*}
    \pr{\abs{\randSetOne\cap\randSetTwo} \ge 2 \alpha^2 n} = \sum_{S \in \binom{[n]}{\alpha n}} \pr{\randSetOne = S} \pr{\abs{\randSetOne\cap\randSetTwo} \ge 2 \alpha^2 n \st \randSetOne = S}
    .
  \end{align*}
  Conditional on \(\randSetOne = S\), \(\abs{\randSetOne \cap \randSetTwo}\) is a hypergeometric random variable with
  \begin{align*}
    \expect{\abs{\randSetOne \cap \randSetTwo} \st \randSetOne = S}
     & = \sum_{i \in S} \pr{i \in \randSetTwo}
    \\ & = k \pr{1 \in \randSetTwo}
    \\ & = \frac{k^2}{n}
    \\ & = \alpha^2 n
    .
  \end{align*}
  Since the Chernoff bounds hold also for the hypergeometric distribution \cite[Theorem 1.10.25]{doerrProbabilisticToolsAnalysis2020}, we get that
  \begin{align*}
    \pr{\abs{\randSetOne\cap\randSetTwo} \ge 2 \alpha^2 n \st \randSetOne = S} \le \exp\left[-\frac{\alpha^2 n}{3}\right]
    .
  \end{align*}
  Hence,
  \begin{align}
    \pr{\abs{\randSetOne\cap\randSetTwo} \ge 2 \alpha^2 n}
     & = \sum_{S \in \binom{n}{\alpha n}} \pr{\abs{\randSetOne\cap\randSetTwo} \ge 2 \alpha^2 n \st \randSetOne = S} \pr{\randSetOne = S} \nonumber
    \\ & \le \exp\left[-\frac{\alpha^2 n}{3}\right] \cdot \sum_{S \in \binom{n}{\alpha n}} \pr{\randSetOne = S} \nonumber
    \\ & = \exp\left[-\frac{\alpha^2 n}{3}\right]
    .\label{eq:thmRSSP:term6}
  \end{align}
  Finally, we take care of the term
  \begin{align}
    \frac{\pr{\abs{\randSetOne \cap \randSetTwo} \ge 2\alpha^2 n}}{\pr{\cH_{S_0}}}
    .\label{eq:thmRSSP:term7}
  \end{align}
  By plugging \cref{eq:thmRSSP:term6,eq:denominator_bound}
  in \cref{eq:thmRSSP:term7}, we obtain
  \begin{align}
    \frac{\pr{\abs{\randSetOne \cap \randSetTwo} \ge 2\alpha^2 n}}{\pr{\cH_{S_0}}}
     & \le 256\exp\left[d\log \frac{2\pi \alpha n (1 + 2\sqrt{c_d} + 2c_d)}{4\eps^2} - \frac{\alpha^2 n}{3}\right] \nonumber
    \\ & \le 1 \label{eq:thmRSSP:term8}
    ,
  \end{align}
  which is true whenever \(n \ge a_5 d^2 (\log d + \log \frac 1\eps)\) for some large enough constant \(a_5 > 0\).
  By \cref{eq:thmRSSP:term8,eq:mrssp:small-intersection}, we obtain that
  \begin{align*}
    \pr{T > 0} \ge \delta
    ,
  \end{align*}
  for some positive constant \(\delta > 0\).

\subsection{Proving SLTH for structured pruning}
\label{subsec:proof_slth_kernel}
We first show how to obtain the same approximation result for a single-layer CNN.
Then, we iteratively apply the same argument for all layers of a larger CNN and show that the approximation error stays small.

We define the \emph{positive} and \emph{negative} parts of a tensor.

\begin{definition}
  \label{def:posnegpart}Given a tensor $\tX \in \R^{d_1\times \ldots \times d_{n}}$, the \emph{positive} and
  \emph{negative} parts \(\tX^+\) and \(\tX^-\) of $\tX$ are respectively defined as $\tX^+_{\vec{i}} = \tX_{\vec{i}} \cdot \ind_{\tX_{\vec{i}} > 0}$
  and $\tX^-_{\vec{i}} = -\tX_{\vec{i}} \cdot \ind_{\tX_{\vec{i}} < 0}$,
  where $\vec{i} \in [d_1] \times \ldots \times [d_n]$ points at a generic entry of $\tX$.
\end{definition}

Let us define some convenient notations before proceeding with the proof.
By \([n:m]\) we denote the set \(\{n, n + 1, \ldots, m\}\) for each pair of integers \(n \le m \in \N\).

\subsubsection*{Approximating a single-layer CNN}

We first present a preliminary lemma that shows how to prune a single-layer convolution $\relu\left(\tV \ast \tX\right)$
in a way that dispenses us from dealing with the ReLU $\relu$.

\begin{lemma}
  \label{lemma:droprelu}Let $D, d, c, n \in \N$ be positive
  integers,
  $\tV \in \R^{1\times 1\times c\times 2n c}$,
  and $\tX \in \R^{D\times D\times c}$.
  Let $S_1 \in \left\{0, 1\right\} ^{\size\left(\tV\right)}$ be a \(2n\)-channel-blocked mask.
  There exists a mask \(S_2 \in \left\{0, 1\right\} ^{\size\left(\tV\right)}\) which only removes filters such that, for each $\left(i, j, k\right) \in \left[D\right]\times \left[D\right]\times \left[2n c\right]$, if \(\widetilde{\sS} = S_1 \ast S_2\), then
  \[
    \Bigl(\relu\Bigl(\Bigl(\tV\ast\widetilde{\sS}\Bigr)\ast \tX\Bigr)\Bigr)_{i, j, k} = \Bigl(\Bigl(\tV\ast\widetilde{\sS}\Bigr)^{+}\ast \tX^{+} + \Bigl(\tV\ast\widetilde{\sS}\Bigr)^{-}\ast \tX^{-}\Bigr)_{i, j, k}
    .
  \]
\end{lemma}

\begin{proof}
  $\sS_2 \in \left\{0, 1\right\} ^{\size\left(\tV\right)}$ is such that $\widetilde{\tV} = \tV\ast\widetilde{\sS} = (\tV\ast \sS_1) \ast \sS_2$ contains only non-negative edges going
  from each input channel $t$ to the output channels $2(t-1)n + 1, \ldots, (2t -1)n$,
  and only non-positive
  edges going from each input channel $t$ to the output channels $(2t -1)n + 1, \ldots, 2tn$,
  while all remaining edges are set to zero.\footnote{We consider $0$ to be both non-negative and non-positive.}
  In formulas, we obtain
  a tensor $\widetilde{\tV} = \tV\ast\widetilde{\sS}$ such that, for each $\left(t, k\right) \in \left[c\right]
  \times \left[2n c\right]$:
  \begin{align}
    \left(\tV\ast\widetilde{\sS}\right)_{1, 1, t, k} = \begin{cases}
      \tV_{1, 1, t, k} \cdot \ind_{\tV_{1, 1, t, k} > 0} &\text{if }k \in \left[\left(2t-2\right)n + 1:\left(2t-1\right)n\right], \\
      \tV_{1, 1, t, k} \cdot \ind_{\tV_{1, 1, t, k} < 0} &\text{if }k \in \left[\left(2t-1\right)n + 1:2tn\right], \\
      0 &\text{otherwise.}
    \end{cases}\label{eq:Vstar}
  \end{align}
  To simplify the notation, we define the following indicator functions: for any $\left(t, k\right) \in \left[c\right]\times \left[2n c\right]$,
  \begin{align}
    \ind_{\frac{k}{2n} \in \left(t-1, t-\frac{1}{2}\right]} = 1 \text{ iff } k \in \left[\left(2t-2\right)n + 1:\left(2t-1\right)n\right], \text{and}\nonumber \\
    \ind_{\frac{k}{2n} \in \left(t-\frac{1}{2}, t\right]} = 1 \text{ iff } k \in \left[\left(2t-1\right)n + 1:2tn\right]
    .\label{eq:k_vs_t}
  \end{align}
  For each $\left(i, j, k\right) \in \left[D\right]\times \left[D\right]\times \left[2n c\right]$, applying \cref{eq:Vstar} and using \cref{def:posnegpart},
  it then holds that
  \begin{align}
    & \left(\relu\left(\left(\tV\ast\widetilde{\sS}\right)\ast \tX\right)\right)_{i, j, k} \nonumber
    \\ = \ & \relu\left(\sum_{t = 1}^{c_0}\widetilde{\tV}_{1, 1, t, k}\tX_{i, j, t}\right)\nonumber
    \\ = \ & \relu\bigg(\sum_{t = 1}^{c_0}\big(\tV_{1, 1, t, k}\tX_{i, j, t} \cdot \ind_{\tV_{1, 1, t, k} > 0}\ind_{\frac{k}{2n} \in \left(t-1, t-\frac{1}{2}\right]} \nonumber + V_{1, 1, t, k}\tX_{i, j, t} \cdot \ind_{\tV_{1, 1, t, k} < 0}\ind_{\frac{k}{2n} \in \left(t-\frac{1}{2}, t\right]}\big)\bigg) \nonumber
    \\ = \ & \relu\left(\sum_{t = 1}^{c_0}\left({\tV}_{1, 1, t, k}^{+}\tX_{i, j, t}\ind_{\frac{k}{2n} \in \left(t-1, t-\frac{1}{2}\right]}-{\tV}_{1, 1, t, k}^{-}\tX_{i, j, t}\ind_{\frac{k}{2n} \in \left(t-\frac{1}{2}, t\right]}\right)\right) \nonumber
    \\ = \ & \relu\bigg(\sum_{t = 1}^{c_0}\big({\tV}_{1, 1, t, k}^{+}(\tX_{i, j, t}^{+} - \tX_{i, j, t}^{-})\ind_{\frac{k}{2n} \in \left(t-1, t-\frac{1}{2}\right]} + {\tV}_{1, 1, t, k}^{-}(\tX_{i, j, t}^{-} - \tX_{i, j, t}^{+})\ind_{\frac{k}{2n} \in \left(t-\frac{1}{2}, t\right]}\big)\bigg)
    .\label{eq:relu-trick1}
  \end{align}
  Observe that only one term survives in the summation in \cref{eq:relu-trick1}, as there exists only one \(t \in [c_0]\) such that \(k \in [(2t-2)n + 1: 2tn]\), say \(t^\star\).
  Moreover, out of the four additive terms in the expression
  \[
    {\tV}_{1, 1, t^\star, k}^{+}(\tX_{i, j, t^\star}^{+} - \tX_{i, j, t^\star}^{-})\ind_{\frac{k}{2n} \in \left(t^\star-1, t^\star-\frac{1}{2}\right]} + {\tV}_{1, 1, t^\star, k}^{-}(\tX_{i, j, t^\star}^{-} - \tX_{i, j, t^\star}^{+})\ind_{\frac{k}{2n} \in \left(t^\star-\frac{1}{2}, t^\star\right]}
    ,
  \]
  at most one is non-zero, due to \cref{def:posnegpart}.
  The ReLU cancels out negative ones, implying that \cref{eq:relu-trick1} can be rewritten without the ReLU as a sum of only non-negative terms (out of which, at most one is non-zero) as follows
  \begin{align}
    & \relu\bigg(\sum_{t = 1}^{c_0}\big({\tV}_{1, 1, t, k}^{+}(\tX_{i, j, t}^{+} - \tX_{i, j, t}^{-})\ind_{\frac{k}{2n} \in \left(t-1, t-\frac{1}{2}\right]} \nonumber + {\tV}_{1, 1, t, k}^{-}(\tX_{i, j, t}^{-} - \tX_{i, j, t}^{+})\ind_{\frac{k}{2n} \in \left(t-\frac{1}{2}, t\right]}\big)\bigg) \nonumber
    \\ = \ & \sum_{t = 1}^{c_0}\left({\tV}_{1, 1, t, k}^{+}\tX_{i, j, t}^{+}\ind_{\frac{k}{2n} \in \left(t-1, t-\frac{1}{2}\right]} + {\tV}_{1, 1, t, k}^{-}\tX_{i, j, t}^{-}\ind_{\frac{k}{2n} \in \left(t-\frac{1}{2}, t\right]}\right) \label{eq:big_ind_functions_trick}
    .
  \end{align}
  Finally, by \cref{eq:Vstar} and \cref{eq:k_vs_t}, ${{\widetilde{\tV}}_{1, 1, t, k}^{+} = 0}$
  if ${\frac{k}{2n}\not \in \left(t-1, t-\frac{1}{2}\right]}$,
  and ${{\widetilde{\tV}}_{1, 1, t, k}^{-} = 0}$ if ${\frac{k}{2n} \in \left(t-\frac{1}{2}, t\right]}$,
  which means that in \cref{eq:big_ind_functions_trick} we can ignore
  the indicator functions and further simplify the expression as
  \begin{align*}
    & \sum_{t = 1}^{c_0}\left({\tV}_{1, 1, t, k}^{+}\tX_{i, j, t}^{+}\ind_{\frac{k}{2n} \in \left(t-1, t-\frac{1}{2}\right]} + {\tV}_{1, 1, t, k}^{-}\tX_{i, j, t}^{-}\ind_{\frac{k}{2n} \in \left(t-\frac{1}{2}, t\right]}\right)
    \\ = \ & \sum_{t = 1}^{c_0}\left({\widetilde{\tV}}_{1, 1, t, k}^{+}\tX_{i, j, t}^{+} + {\widetilde{\tV}}_{1, 1, t, k}^{-}\tX_{i, j, t}^{-}\right)
    \\ = \ & \left(\sum_{t = 1}^{c_0}{\widetilde{\tV}}_{1, 1, t, k}^{+}\tX_{i, j, t}^{+} + \sum_{t = 1}^{c_0}{\widetilde{\tV}}_{1, 1, t, k}^{-}\tX_{i, j, t}^{-}\right)
    \\ = \ & \left({\widetilde{\tV}}^{+}\ast \tX^{+} + {\widetilde{\tV}}^{-}\ast \tX^{-}\right)_{i, j, k}
    ,
  \end{align*}
  yielding the desired result.
\end{proof}

We approximate a single convolution \(\tK \ast \tX\) by pruning a polynomially larger neural network of the form \(\tU \ast \relu(\tL \ast \tX)\) exploiting only a channel-blocked mask and filter removal:
this is achieved using the MRSS result (\cref{thm:mrrs_gaussian_scaled}).

\begin{lemma}[Kernel pruning]
  \label{lemma:kernelpruning}
  Let $D, d, c_0, c_1, n \in \N$ be positive
  integers, $\eps \in \left(0, \frac{1}{4}\right), M \in \R_{>0}$, and
  $C \in \R_{>0}$ be universal constants, with
  \[
    n \ge Cd^{5}c_1^{5}\log^{2}\frac{d c_1c_0}{\eps}
    .
  \]
  Let $\tU\sim\normal{\sqrt{d}\times \sqrt{d}\times 2nc_0\times c_1}$,
  $\tV\sim\normal{1\times 1\times c_0\times 2nc_0}$ and
  $\sS \in \left\{0, 1\right\} ^{\size\left(\tL\right)}$, with \(\sS\) being a \(2n\)-channel-blocked mask.
  We define $N_{0}\left(\tX\right) = \tU\ast\relu\left(\tV\ast \tX\right)$
  where $\tX \in \R^{D\times D\times c_0}$,
  and its pruned version $N_{0}^{\left(\sS\right)}\left(\tX\right) = \tU\ast\relu\left(\left(\tV\ast \sS\right)\ast \tX\right)$.
  With probability $1 - \eps$, for all $\tK \in \R^{\sqrt{d}\times \sqrt{d}\times c_{0}\times c_1}$
  with $\normone*{\tK_{:, :, t_0, :}} \le \sqrt{k}$, with \(k = n/(6\sqrt{d c_1})\) for all \(t_0 \in [c_0]\),
  it is possible to remove filters from \(N_{0}^{\left(\sS\right)}\) to obtain a CNN \(\widetilde{N}_{0}^{\left(\sS\right)}\) for which
  \[
    \sup_{\tX:\normmax* \tX \le M}\normmax*{\tK\ast \tX-\widetilde{N}_{0}^{\left(\sS\right)}\left(\tX\right)}
    < \eps M
    .
  \]
\end{lemma}
\begin{proof}
  Adopting the same definitions as in \cref{lemma:droprelu} (and Eq.
  \cref{eq:Vstar}), for each $\left(r, s, t_1\right) \in [\sqrt{d}]\times [\sqrt{d}]\times \left[c_1\right]$
  we have, by \cref{lemma:droprelu},
  \begin{align*}
    &\left(\tU*\relu\left(\left(V\ast \widetilde{\sS}\right)\ast \tX\right)\right)_{r, s, t_1}
    \\ = \ & \left(\tU\ast \left(\left(\widetilde{\tV}^{+}\ast \tX^{+}\right) + \left(\widetilde{\tV}^{-}\ast \tX^{-}\right)\right)\right)_{r, s, t_1}
    \\ = \ & \sum_{i, j \in [\sqrt{d}], k \in \left[2nc_0\right]}\tU_{i, j, k, t_1} \cdot \left(\left(\widetilde{\tV}^{+}\ast \tX^{+}\right) + \left(\widetilde{\tV}^{-}\ast \tX^{-}\right)\right)_{r-i + 1, s-j + 1, k}
    \\ = \ & \sum_{i, j \in [\sqrt{d}], k \in \left[2nc_0\right]}\tU_{i, j, k, t_1} \cdot \sum_{t_0 \in \left[c_0\right]} \bigg(\widetilde{\tV}_{1, 1, t_0, k}^{+} \cdot \tX^{+}_{r-i + 1, s-j + 1, t_0} + \widetilde{\tV}_{1, 1, t_0, k}^{-} \cdot \tX^{-}_{r-i + 1, s-j + 1, t_0}\bigg)
    \\ = \ & \sum_{t_0 \in \left[c_0\right]}\sum_{i, j \in [\sqrt{d}], k \in \left[2nc_0\right]}\left(\tU_{i, j, k, t_1} \cdot \widetilde{\tV}_{1, 1, t_0, k}^{+}\right) \cdot \tX^{+}_{r-i + 1, s-j + 1, t_0}
    \\ & + \sum_{t_0 \in \left[c_0\right]}\sum_{i, j \in [\sqrt{d}], k \in \left[2nc_0\right]}\left(\tU_{i, j, k, t_1} \cdot \widetilde{\tV}_{1, 1, t_0, k}^{-}\right) \cdot \tX^{-}_{r-i + 1, s-j + 1, t_0}
    \\ = \ & \sum_{i, j \in [\sqrt{d}], t_0 \in \left[c_0\right]}\left(\sum_{k \in \left[2nc_0\right]}\tU_{i, j, k, t_1} \cdot \widetilde{\tV}_{1, 1, t_0, k}^{+}\right) \cdot \tX^{+}_{r-i + 1, s-j + 1, t_0}
    \\ & + \sum_{i, j \in [\sqrt{d}], t_0 \in \left[c_0\right]}\left(\sum_{k \in \left[2nc_0\right]}\tU_{i, j, k, t_1} \cdot \widetilde{\tV}_{1, 1, t_0, k}^{-}\right) \cdot \tX^{-}_{r-i + 1, s-j + 1, t_0}
    .
  \end{align*}
  We remind the reader that \(\widetilde{\sS} = \sS_1 * \sS_2\), with \(\sS_1\) being a \(2n\)-channel-blocked mask and \(\sS_2\) being a mask that removes filters.
  Define \(\tL^{+}_{i, j, t_0, t_1} = \sum_{k \in \left[2nc_0\right]}\tU_{i, j, k, t_1} \cdot \widetilde{\tV}_{1, 1, t_0, k}^{+} \) and, similarly, \(\tL^{-}_{i, j, t_0, t_1} = \sum_{k \in \left[2nc_0\right]}\tU_{i, j, k, t_1} \cdot \widetilde{\tV}_{1, 1, t_0, k}^{-}\).
  Then,
  \begin{align*}
    & \sum_{i, j \in [\sqrt{d}], t_0 \in \left[c_0\right]}\left(\sum_{k \in \left[2nc_0\right]}\tU_{i, j, k, t_1} \cdot \widetilde{\tV}_{1, 1, t_0, k}^{+}\right) \cdot \tX^{+}_{r-i + 1, s-j + 1, t_0}
    \\ & + \sum_{i, j \in [\sqrt{d}], t_0 \in \left[c_0\right]}\left(\sum_{k \in \left[2nc_0\right]}\tU_{i, j, k, t_1} \cdot \widetilde{\tV}_{1, 1, t_0, k}^{-}\right) \cdot \tX^{-}_{r-i + 1, s-j + 1, t_0}
    \\ = \ & \sum_{i, j \in [\sqrt{d}], t_0 \in \left[c_0\right]}\tL^{+}_{i, j, t_0, t_1} \cdot \tX^{+}_{r-i + 1, s-j + 1, t_0} + \sum_{i, j \in [\sqrt{d}], t_0 \in \left[c_0\right]}\tL^{-}_{i, j, t_0, t_1} \cdot \tX^{-}_{r-i + 1, s-j + 1, t_0}
    .
  \end{align*}

  We now show that, for each $t_0 \in \left[c_0\right]$, $\tK_{:, :, t_0, :}$
  can be $\eps$-approximated by $\tL^{+}_{:, :, t_0, :}$ by suitably
  pruning $\widetilde{\tV}^{+}$, i.e., by further zeroing entries of $\widetilde{\sS}$,
  and that such pruning corresponds to solving an instance of MRSS according
  to \cref{thm:mrrs_gaussian_scaled}.
  The same reasoning applies to $\tK^{-}$ and $\tL^{-}$.

  For each $t_0 \in \left[c_0\right]$, let
  \[
    I_{+}^{(t_0)} = \left\{k \in \left\{\left(2t_0 - 2\right)n + 1, \ldots, \left(2t_0 - 1\right)n\right\} :\widetilde{\sS}_{1, 1, t_0, k} = 1\right\} .
  \]
  Observe that $I_{+}^{(t_0)}$ consists of the strictly positive entries of $\widetilde{\tV}_{1, 1, t_0, :}^{+}$.\footnote{Notice that excluding zero entries implies conditioning on the event that the entry is not zero.
  However, such an event has zero probability and, thus, does not impact the analysis.}
  Since the entries of $\tV$ follow a standard normal distribution, each entry is positive with probability $1/2$.
  By a standard application of Chernoff bounds (\cref{lemma:chernoff-hoeffding} in \cref{app:supporting-results}), we then have
  \begin{align}
    \Pr\left(\abs{I_{+}^{(t_0)}}
    > \frac{n}{3}\right) \ge 1 - \frac{\eps}{4}
    ,\label{eq:lower_on_Iplus}
  \end{align}
  provided that the constant $C$ in the bound on $n$ is sufficiently
  large.

  For each $k \in I_{+}^{(t_0)}$, up to reshaping the tensor as a one-dimensional
  vector, $\tU_{:, :, k, :} \cdot \widetilde{\tV}_{1, 1, t_0, k}^{+}$
  is an NSN vector according to \cref{def:nsn_vector} by \cref{lemma:halfnormalNSN} (\cref{app:supporting-results}).
  Thus, for each
  $t_0 \in \left[c_0\right]$ and a sufficiently-large value of $C$,
  since we have $n \ge Cd^{5}c_1^{5}\log^2\frac{d c_1c_0}{\eps}$,
  we can apply an amplified version of \cref{thm:mrrs_gaussian_scaled}
  (i.e., \cref{cor:mrrs_NSN} in \cref{app:supporting-results} with vectors of dimension $d c_{1}$)
  to show that, with probability $1 - \frac{\eps}{4c_0}$, for all target filter $\tK$ such that $\normone{\tK_{:, :, t_0, :}} \le \sqrt{k}$, with \(k = n/(6\sqrt{d c_1})\), there exists
  a way to zero the entries indexed by $I_{+}^{(t_0)}$ of $\widetilde{\sS}$ (and thus $\widetilde{\tV}_{1, 1, t_0, :}^{+}$),
  so that the pruned version of $\tL^{+}_{:, :, t_0, :} = \sum_{k \in \left[2nc_0\right]}\tU_{:, :, k, :} \cdot \widetilde{\tV}_{1, 1, t_0, k}^{+}$ approximates \(\tK_{:, :, t_0, :}\).
  In particular, there exists another binary mask
  \(\sS_3^{+} \in \{0, 1\}^{\size{\widetilde{S}}}\) such that \(\hat{L}_{:, :, t_0, :}^{+} = \sum_{k \in \left[2nc_0\right]}\tU_{:, :, k, :} \cdot \hat{\tV}_{1, 1, t_0, k}^{+}\) approximates \(\tK_{:, :, t_0, :}\), where \(\hat{\tV}^{+} = \widetilde{\tV}^{+} * \sS_3^{+} \).
  An analogous argument carries on for a binary mask \(\sS_3^{-}\) and \(-\hat{L}_{:, :, t_0, :}^{-}\).\footnote{The negative sign in front of \(\hat{L}_{:, :, t_0, :}^{-}\) does not affect the random subset sum result as each entry is independently negative or positive with the same probability.}
  More formally, let
  \begin{align*}
    \eventkernel_{t_0, +} & = \left\{\forall \tK: \normone*{\tK} \le 1, \exists \sS_3^{+} \in \{0, 1\}^{\size{\widetilde{S}}} \st \normmax*{\hat{L}^{+}_{:, :, t_0, :}-\tK_{:, :, t_0, :}} \le \frac{\eps}{2d c_1c_0}\right\}, \\
    \eventkernel_{t_0, -} & = \left\{\forall \tK: \normone*{\tK} \le 1, \exists \sS_3^{-} \in \{0, 1\}^{\size{\widetilde{S}}} \st \normmax*{\hat{L}^{-}_{:, :, t_0, :} + \tK_{:, :, t_0, :}} \le \frac{\eps}{2d c_1c_0}\right\}, \text{and}\\
    \eventkernel & = \left(\bigcap_{t_0 \in \left[c_0\right]}\eventkernel_{t_0, +}\right)\bigcap\left(\bigcap_{t_0 \in \left[c_0\right]}\eventkernel_{t_0, -}\right)
    .
  \end{align*}
  Then, by \cref{cor:mrrs_NSN},
  \begin{align*}
    \Pr\left(\eventkernel_{t_0, +} \st \abs{I_{+}^{(t_0)}} > \frac{n}{3}\right) \ge 1 - \frac{\eps}{4c_0}, \text{and} \\
    \Pr\left(\eventkernel_{t_0, -} \st \abs{I_{-}^{(t_0)}} > \frac{n}{3}\right) \ge 1 - \frac{\eps}{4c_0}
    .
  \end{align*}
  By the union bound, we have the following:
  \begin{align*}
    & \Pr\left(\eventkernel \st \abs{I_{+}^{(t_0)}}, \abs{I_{-}^{(t_0)}} > \frac{n}{3}\right)
    \\ = \ & 1 - \Pr\left(\left(\bigcup_{t_0 \in \left[c_0\right]}\eventkernel_{t_0, +}\right)\bigcup\left(\bigcup_{t_0 \in \left[c_0\right]}\eventkernel_{t_0, -}\right) \st \min\{\abs{I_{+}^{(t_0)}}, \abs{I_{-}^{(t_0)}}\} > \frac{n}{3}\right)
    \\ \ge \ & 1 - \sum_{t_0 \in \left[c_0\right]}\left[\Pr\left(\eventkernel_{t_0, +} \st \min\{\abs{I_{+}^{(t_0)}}, \abs{I_{-}^{(t_0)}}\} > \frac{n}{3}\right) + \Pr\left(\eventkernel_{t_0, -} \st \min\{\abs{I_{+}^{(t_0)}}, \abs{I_{-}^{(t_0)}}\} > \frac{n}{3}\right)\right]
    \\ \ge \ & 1 - 2 \sum_{t_0 \in \left[c_0\right]}\frac{\eps}{4c_0}
    \\ \ge \ & 1 - \frac{\eps}{2}
    .
  \end{align*}
  Since $\Pr\left(\min\{\abs{I_{+}^{(t_0)}}, \abs{I_{-}^{(t_0)}}\} > \frac{n}{3}\right) \ge 1 - \frac{\eps}{2}$, we can remove the conditional event obtaining
  \begin{align}
    & \Pr\left(\eventkernel\right)
    \\ \ge \ & \Pr\left(\eventkernel \st \min\{\abs{I_{+}^{(t_0)}}, \abs{I_{-}^{(t_0)}}\} > \frac{n}{3}\right)\Pr\left(\min\{\abs{I_{+}^{(t_0)}}, \abs{I_{-}^{(t_0)}}\} > \frac{n}{3}\right)\nonumber
    \\ \ge \ & \left(1 - \frac{\eps}{2}\right)^{2}\nonumber
    \\ \ge \ & 1 - \eps
    .\label{eq:prob_filter_apx}
  \end{align}
  To rewrite the latter in terms of the
  filter $\tK$ and a mask $\hat{\sS}$, we notice that pruning \({L}^{+}_{:, :, t_0, :}\) and \({L}^{-}_{:, :, t_0, :}\) separately, with two binary masks, is equivalent to say that there exists a single binary mask \(\sS_3 \in \{0, 1\}^{\size{\widetilde{S}}}\) such that, \(\hat{L}_{:, :, t_0, :}\) can be written as \(\hat{L}_{:, :, t_0, :} = \sum_{k \in \left[2nc_0\right]}\tU_{:, :, k, :} \cdot \hat{\tV}_{1, 1, t_0, k}\), where \(\hat{\tV} = \widetilde{\tV} * \sS_3 \).
  \cref{eq:prob_filter_apx} implies that, with probability \(1 - \eps\), for all target filters
  $\tK$ such that $\normone{\tK_{:, :, t_0, :}} \le \sqrt{k}$, with \(k = n/(6\sqrt{d c_1})\), such \(\sS_3\) exists and hence,
  \begin{align}
    \normmax*{{\tK-\hat{L}^+}} + \normmax*{{\tK + \hat{L}^-}}
    \le \frac{\eps}{d c_1c_0}
    .\label{eq:bound_on_kernel_apx}
  \end{align}
  Let \(\hat{\sS} = \widetilde{S} * \sS_3\): then
  \(\hat{\sS} = \sS_1 * \sS_2 * \sS_3\) is the combination of a \(2n\)-channel-blocked mask \(S = \sS_1\) and a mask \(\sS_2 * \sS_3\) that only removes filters.

  Also, whenever such binary mask \(\hat{\sS}\) exists, we have that
  \begin{align*}
    & \sup_{\tX:\normmax* \tX \le M}\normmax*{\tK\ast \tX-N_{0}^{\left(\hat{\sS}\right)}\left(\tX\right)}
    \\ = \ & \sup_{\tX:\normmax* \tX \le M}\normmax*{\tK\ast \tX-\tU\ast\relu\left((V* \hat{\sS})\ast \tX\right)}
    \\ = \ & \sup_{\tX:\normmax* \tX \le M}\normmax*{\tK\ast \tX-\tU\ast\relu\left((V* \widetilde{S} * \sS_3)\ast \tX\right)}
    \\ = \ & \sup_{\tX:\normmax* \tX \le M}\normmax*{\tK\ast\left(\tX^{+}-\tX^{-}\right)-\tU\ast\left(\left(\hat{\tV}^{+}\ast \tX^{+}\right) + \left(\hat{\tV}^{-}\ast \tX^{-}\right)\right)}
    ,
  \end{align*}
  where the latter holds by \cref{lemma:droprelu}.\footnote{The presence of \(\sS_3\) does not influence the proof of \cref{lemma:droprelu}.}
  Then, by the distributive property of the convolution and the triangle inequality,
  \begin{align*}
    & \sup_{\tX:\normmax* \tX \le M}\normmax*{\tK\ast\left(\tX^{+}-\tX^{-}\right)-\tU\ast\left(\left(\hat{\tV}^{+}\ast \tX^{+}\right) + \left(\hat{\tV}^{-}\ast \tX^{-}\right)\right)}
    \\ = \ & \sup_{\tX:\normmax* \tX \le M}\normmax*{\tK\ast \tX^{+}-\tU\ast\left(\hat{\tV}^{+}\ast \tX^{+}\right)-\tK\ast \tX^{-}-\tU\ast\left(\hat{\tV}^{-}\ast \tX^{-}\right)}
    \\ \le \ & \sup_{\tX:\normmax* \tX \le M}\normmax*{\tK\ast \tX^{+}-\tU\ast\left(\hat{\tV}^{+}\ast \tX^{+}\right)} + \sup_{\tX:\normmax* \tX \le M}\normmax*{\tK\ast \tX^{-} + \tU\ast\left(\hat{\tV}^{-}\ast \tX^{-}\right)}
    .
  \end{align*}
  One can now apply the Tensor Convolution Inequality (\cref{lemma:convineq}) and obtain
  \begin{align*}
    & \sup_{\tX:\normmax* \tX \le M}\normmax*{\tK\ast \tX^{+}-\tU\ast\left(\hat{\tV}^{+}\ast \tX^{+}\right)} + \sup_{\tX:\normmax* \tX \le M}\normmax*{\tK\ast \tX^{-} + \tU\ast\left(\hat{\tV}^{-}\ast \tX^{-}\right)}
    \\ \le \ & \sup_{\tX:\normmax* \tX \le M}\normmax*{\tX^{+}} \cdot \normone*{\tK-\tU\ast\hat{\tV}^{+}} + \sup_{\tX:\normmax* \tX \le M}\normmax*{\tX^{-}} \cdot \normone*{\tK + \tU\ast\hat{\tV}^{-}}
    \\ = \ & M \cdot \normone*{\tK-\tU\ast\hat{\tV}^{+}} + M \cdot \normone*{\tK + \tU\ast\hat{\tV}^{-}}
    .
  \end{align*}
  Now, observing that the number of entries of the two tensors in the expression above is \(d c_1c_0\), and using \cref{eq:bound_on_kernel_apx}, we get that
  \begin{align*}
    M \cdot \normone*{\tK-\tU\ast\hat{\tV}^{+}} + M \cdot \normone*{\tK-\tU\ast\hat{\tV}^{-}}
     & \le d c_1c_0 \left(\normmax*{\tK-\tU\ast\hat{\tV}^{+}} + \normmax*{\tK-\tU\ast\hat{\tV}^{-}}\right)
    \\ & \le d c_1c_0 M\frac{\eps}{d c_1c_0}
    \\ & = \eps M
    .
  \end{align*}
\end{proof}

\begin{proof}[Proof of \cref{thm:slth_kernel}.]
  To bound the error propagation across layers, we define the layers' outputs
  \begin{align}
    \tX^{(0)} & = \tX, \nonumber \\
    \tX^{(i)} & = \relu\left(\tK^{\left(i\right)}\ast\tX^{(i-1)}\right) \text{for }1 \le i \le \ell
    .\label{eq:intermediate_layer-1}
  \end{align}
  Notice that $\tX^{(\ell)}$ is the output of the target function,
  i.e., $f\left(\tX\right) = \tX^{(\ell)}$.

  For brevity's sake, given masks $\tS^{(2i - 1)} \in \{0, 1\}^{\text{size}(\tL^{(2i - 1)})}$ for each $i = 1, \ldots, \ell$,
  let us denote
  \begin{align}
    \widetilde{\tL}^{(2i - 1)} = \tL^{(2i - 1)}*\tS^{(2i - 1)}
    .\label{eq:pruned_shortcut-1}
  \end{align}
  We now show that the random network approximates the target one after being pruned only at layers $\tL^{(2i - 1)}$ for all $i = 1, \ldots, \ell$, by masks $\tS^{(2i - 1)}$ which are the composition of a $2n_{2i - 1}$-channel-blocked mask and a mask that only removes filters.
  The key idea is to iteratively applying \cref{lemma:kernelpruning}.
  Since the ReLU function is $1$-Lipschitz, for all $i$ it holds that
  \begin{align}
    & \normmax*{\relu\left(\tK^{(i)}\ast\tX^{(i-1)}\right)-\relu\left(\tL^{(2i)}\ast\relu\left(\widetilde{\tL}^{(2i-1)}\ast\tX^{(i-1)}\right)\right)}
    \\ \le \ & \normmax*{\tK^{(i)}\ast\tX^{(i-1)}-\tL^{(2i)}\ast\relu\left(\widetilde{\tL}^{(2i-1)}\ast\tX^{(i-1)}\right)}
    .\label{eq:relu_lipschitz-1}
  \end{align}
  The key step of the proof is that, for each layer $i$, since $n_{i} \ge Cd_i^{5}c_{i}^{5}\log^{2}\frac{2d c_{i}c_{i-1}\ell}{\eps}$
  for a suitable constant $C$, we can apply \cref{lemma:kernelpruning}
  to get that, with probability at least $1 - \frac{\eps}{2\ell}$, for all $i$ and for all choices of target filters
  $\tK^{(i)} \in \R^{\sqrt{d_i} \times \sqrt{d_i} \times c_{i-1} \times c_i}$
  it holds
  \begin{align}
    \normmax*{\tK^{(i)}\ast\tX^{(i-1)}-\tL^{(2i)}\ast\relu\left(\widetilde{\tL}^{(2i-1)}\ast\tX^{(i-1)}\right)}
    < \frac{\eps}{2\ell} \cdot \normmax*{\tX^{(i-1)}}
    .\label{eq:layer_apx_pre-1}
  \end{align}
  Hence, combining \cref{eq:relu_lipschitz-1} and \cref{eq:layer_apx_pre-1}
  we get that, with probability at least $1 - \frac{\eps}{2\ell}$, for all $i$ and all choices of target filters
  $\tK^{(i)} \in \R^{\sqrt{d_i} \times \sqrt{d_i} \times c_{i-1} \times c_i}$ with ${\normone*{\tK^{(i)}_{:, :, t_{i-1}, :}} \le \sqrt{k_i}}$, with \(k_i = n_i/(6\sqrt{d_i c_i})\) for all \(t_{i-1} \in [c_{i-1}]\),
  \begin{align}
    \normmax*{\relu\left(\tK^{(i)}\ast\tX^{(i-1)}\right)-\relu\left(\tL^{(2i)}\ast\relu\left(\widetilde{\tL}^{(2i-1)}\ast\tX^{(i-1)}\right)\right)}
    < \frac{\eps}{2\ell} \cdot \normmax{\tX^{(i-1)}}
    .\label{eq:layer_apx-1}
  \end{align}
  By a union bound, with probability at least $1 - \eps$, we get that \cref{eq:layer_apx-1} holds for
  all layers $i$ and all choices of target filters $\tK^{(i)} \in \R^{\sqrt{d_i} \times \sqrt{d_i} \times c_{i-1} \times c_i}$.

  Analogously, we can define the pruned layers' outputs
  \begin{align}
    \widetilde{\tX}^{(0)} & = \tX, \nonumber \\
    \widetilde{\tX}^{(i)} & = \relu\left(\tL^{(2i)}\ast\relu\left(\widetilde{\tL}^{(2i-1)}\ast\widetilde{\tX}^{(i-1)}\right)\right) \text{for }1 \le i \le \ell
    .\label{eq:intermediate_layer_apx-1}
  \end{align}
  Notice that $\widetilde{\tX}^{(\ell)}$ is the output of the
  pruned network, i.e., $N_{0}^{(S^{(1)}, \ldots, S^{(2\ell)})}(\tX) = \widetilde{\tX}^{(\ell)}.$

  By the same reasoning employed to derive \cref{eq:layer_apx_pre-1}
  and \cref{eq:layer_apx-1} we have that, with probability $1 - \eps$, for all layers $i$ and all choices of target filters $\tK^{(i)} \in \R^{\sqrt{d_i} \times \sqrt{d_i} \times c_{i-1} \times c_i}$,
  the output of all pruned layers satisfies
  \begin{align}
    \normmax*{\relu\left(\tK^{(i)}\ast\widetilde{\tX}^{(i-1)}\right)-\relu\left(\tL^{(2i)}\ast\relu\left(\widetilde{\tL}^{(2i-1)}\ast\widetilde{\tX}^{(i-1)}\right)\right)}
    < \frac{\eps}{2\ell} \cdot \normmax*{\widetilde{\tX}^{(i-1)}}
    .\label{eq:rec_apx_out-1}
  \end{align}
  Moreover, for each $1 \le i \le \ell-1$, by the triangle inequality and by \cref{eq:rec_apx_out-1},
  \begin{align*}
    \normmax*{\widetilde{\tX}^{(i)}}
     & = \normmax*{\widetilde{\tX}^{(i)}-\relu\left(\tK^{(i)}\ast\widetilde{\tX}^{(i-1)}\right) + \relu\left(\tK^{(i)}\ast\widetilde{\tX}^{(i-1)}\right)}
    \\ & \le  \normmax*{\widetilde{\tX}^{(i)}-\relu\left(\tK^{(i)}\ast\widetilde{\tX}^{(i-1)}\right)} + \normmax*{\relu\left(\tK^{(i)}\ast\widetilde{\tX}^{(i-1)}\right)}
    \\ & \le  \frac{\eps}{2\ell} \cdot \normmax*{\widetilde{\tX}^{(i-1)}} + \normmax*{\relu\left(\tK^{(i)}\ast\widetilde{\tX}^{(i-1)}\right)}
    .
  \end{align*}
  By the Lipschitz property of \(\relu\) and \cref{lemma:convineq},
  \begin{align*}
    \frac{\eps}{2\ell} \cdot \normmax*{\widetilde{\tX}^{(i-1)}} + \normmax*{\relu\left(\tK^{(i)}\ast\widetilde{\tX}^{(i-1)}\right)}
     & \le \frac{\eps}{2\ell} \cdot \normmax*{\widetilde{\tX}^{(i-1)}} + \normmax*{\tK^{(i)}\ast\widetilde{\tX}^{(i-1)}}
    \\ & \le \frac{\eps}{2\ell} \cdot \normmax*{\widetilde{\tX}^{(i-1)}} + \normone*{\tK^{(i)}}\normmax*{\widetilde{\tX}^{(i-1)}}
    \\ & = \normmax*{\widetilde{\tX}^{(i-1)}}\left(1 + \frac{\eps}{2\ell}\right)
    .
  \end{align*}
  When we unroll the recurrence, we get that, with probability \(1 - \eps\), for all choices of target filters $\tK^{(i)} \in \R^{\sqrt{d_i} \times \sqrt{d_i} \times c_{i-1} \times c_i}$,
  \begin{align}
    \normmax*{\widetilde{\tX}^{(i)}} \le \normmax*{\widetilde{\tX}^{(0)}}\left(1 + \frac{\eps}{2\ell}\right)^{i}
    .\label{eq:unrolling_apx_bound-1}
  \end{align}
  Thus, combining \cref{eq:rec_apx_out-1} and \cref{eq:unrolling_apx_bound-1}, with probability \(1 - \eps\) we get that, for each $i \in \left[\ell\right]$ and for all choices of target filters $\tK^{(i)} \in \R^{\sqrt{d_i} \times \sqrt{d_i} \times c_{i-1} \times c_i}$,
  \begin{align}
    \normmax*{\tK^{(i)}\ast\widetilde{\tX}^{(i-1)}-\tL^{(2i)}\ast\relu\left(\widetilde{\tL}^{(2i-1)}\ast\widetilde{\tX}^{(i-1)}\right)} < \frac{\eps}{2\ell} \cdot \left(1 + \frac{\eps}{2\ell}\right)^{i-1}\normmax*{\widetilde{\tX}^{(0)}}
    .\label{eq:interm_layer_cumul-1}
  \end{align}

  We then see that, with probability $1 - \eps$, for $1 \le i \le \ell$
  and for all choices of target filters $\tK^{(i)} \in \R^{\sqrt{d_i} \times \sqrt{d_i} \times c_{i-1} \times c_i}$, by \cref{eq:intermediate_layer-1} and \cref{eq:intermediate_layer_apx-1}, and by the triangle inequality,
  \begin{align*}
    \normmax*{\tX^{(\ell)}-\widetilde{\tX}^{(\ell)}}
     & = \normmax*{\relu\left(\tK^{\left(\ell\right)}\ast\tX^{(\ell-1)}\right)-\relu\left(\tL^{(2\ell)}\ast\relu\left(\widetilde{\tL}^{(2\ell-1)}\ast\widetilde{\tX}^{(\ell-1)}\right)\right)}
    \\ & \le \normmax*{\relu({\tK^{\left(\ell\right)}\ast\tX^{(\ell-1)}})-\relu(({\tK^{\left(\ell\right)}\ast\widetilde{\tX}^{(\ell-1)}}))}
    \\ &  + \normmax*{\relu({\tK^{\left(\ell\right)}\ast\widetilde{\tX}^{(\ell-1)}})-\relu({\tL^{(2\ell)}\ast\relu\left(\widetilde{\tL}^{(2\ell-1)}\ast\widetilde{\tX}^{(\ell-1)}\right)})}
    .
  \end{align*}
  Again by the $1$-Lipschitz property of the ReLU activation function, and by the distributive property of the convolution operation,
  \begin{align*}
    & \normmax*{\relu({\tK^{\left(\ell\right)}\ast\tX^{(\ell-1)}})-\relu({\tK^{\left(\ell\right)}\ast\widetilde{\tX}^{(\ell-1)}})} + \normmax*{\relu({\tK^{\left(\ell\right)}\ast\widetilde{\tX}^{(\ell-1)}})-\relu({\tL^{(2\ell)}\ast\relu\left(\widetilde{\tL}^{(2\ell-1)}\ast\widetilde{\tX}^{(\ell-1)}\right)})}
    \\ \le \ & \normmax*{\tK^{\left(\ell\right)}\ast\tX^{(\ell-1)}-\tK^{\left(\ell\right)}\ast\widetilde{\tX}^{(\ell-1)}}
    \\ & + \normmax*{\tK^{\left(\ell\right)}\ast\widetilde{\tX}^{(\ell-1)}-\tL^{(2\ell)}\ast\relu\left(\widetilde{\tL}^{(2\ell-1)}\ast\widetilde{\tX}^{(\ell-1)}\right)}
    \\ = \ & \normmax*{\tK^{\left(\ell\right)}\ast\left(\tX^{(\ell-1)}-\widetilde{\tX}^{(\ell-1)}\right)}
    \\ & + \normmax*{\tK^{\left(\ell\right)}\ast\widetilde{\tX}^{(\ell-1)}-\tL^{(2\ell)}\ast\relu\left(\widetilde{\tL}^{(2\ell-1)}\ast\widetilde{\tX}^{(\ell-1)}\right)}
    .
  \end{align*}
  \cref{lemma:convineq} and the hypothesis \(\normone*{\tK^{\left(\ell\right)}} \le 1\) imply that
  \begin{align*}
    \normmax*{\tK^{\left(\ell\right)}\ast\left(\tX^{(\ell-1)}-\widetilde{\tX}^{(\ell-1)}\right)}
    & + \normmax*{\tK^{\left(\ell\right)}\ast\widetilde{\tX}^{(\ell-1)}-\tL^{(2\ell)}\ast\relu\left(\widetilde{\tL}^{(2\ell-1)}\ast\widetilde{\tX}^{(\ell-1)}\right)}
    \\ &  \le \normone*{\tK^{\left(\ell\right)}} \cdot \normmax*{\left(\tX^{(\ell-1)}-\widetilde{\tX}^{(\ell-1)}\right)}
    \\ & \ \ + \normmax*{\tK^{\left(\ell\right)}\ast\widetilde{\tX}^{(\ell-1)}-\tL^{(2\ell)}\ast\relu\left(\widetilde{\tL}^{(2\ell-1)}\ast\widetilde{\tX}^{(\ell-1)}\right)}
    \\ &  \le M \normmax*{\left(\tX^{(\ell-1)}-\widetilde{\tX}^{(\ell-1)}\right)}
    \\ & \ \ + \normmax*{\tK^{\left(\ell\right)}\ast\widetilde{\tX}^{(\ell-1)}-\tL^{(2\ell)}\ast\relu\left(\widetilde{\tL}^{(2\ell-1)}\ast\widetilde{\tX}^{(\ell-1)}\right)}
    .
  \end{align*}
  Now, we first apply \cref{eq:interm_layer_cumul-1} and then we unroll the recurrence for all layers (as, with probability \(1 - \eps\), \cref{eq:interm_layer_cumul-1} holds for all layers and for all choices of target filters $\tK^{(i)} \in \R^{\sqrt{d_i} \times \sqrt{d_i} \times c_{i-1} \times c_i}$), obtaining
  \begin{align*}
    M \normmax*{\left(\tX^{(\ell-1)}-\widetilde{\tX}^{(\ell-1)}\right)}
    & + \normmax*{\tK^{\left(\ell\right)}\ast\widetilde{\tX}^{(\ell-1)}-\tL^{(2\ell)}\ast\relu\left(\widetilde{\tL}^{(2\ell-1)}\ast\widetilde{\tX}^{(\ell-1)}\right)}
    \\ &  \le M \normmax*{\tX^{(\ell-1)}-\widetilde{\tX}^{(\ell-1)}} + \frac{\eps}{2\ell} \cdot \left(1 + \frac{\eps}{2\ell}\right)^{\ell-1}
    \\ &  \le \sum_{j = 1}^{\ell}\frac{\eps}{2\ell} \cdot \left(1 + \frac{\eps}{2\ell}\right)^{j-1}(\prod_{i = 2}^{j-1} M)
    \\ &  \le \sum_{j = 1}^{\ell}\frac{\eps}{2\ell} \cdot \left(1 + \frac{\eps}{2\ell}\right)^{j-1}
  \end{align*}
  By summing the geometric series and observing that \(\eps < 1\), we conclude that
  \begin{align*}
    \sum_{j = 1}^{\ell}\frac{\eps}{2\ell} \cdot \left(1 + \frac{\eps}{2\ell}\right)^{j-1} & = \left(1 + \frac{\eps}{2\ell}\right)^{\ell}-1
    \\ & \le e^{\frac{\eps}{2}} - 1
    \\ & \le \eps
    .
  \end{align*}
  Hence,
  with probability \(1 - \eps\), for all choices of target filters $\tK^{(i)} \in \R^{\sqrt{d_i} \times \sqrt{d_i} \times c_{i-1} \times c_i}$ and  all \(\tX \in [-1, 1]^{D\times D\times c_0}\), for all \(\ell \in [c]\) it holds that
  \[
    \normmax{\tX^{(\ell)}-\widetilde{\tX}^{(\ell)}} \le \eps
    .
  \]
\end{proof}
\section{Limitations and future work}
\label{sec:limitations}
In previous works \citep{daCunhaNV22,Burkholz22} the assumption that the kernel of every second layer has shape $1 \times 1 \times \ldots$ is only an artifact of the proof since one can readily prune entries of an arbitrarily shaped tensor to enforce the desired shape.
In our case, however, the concept of structured pruning can be quite broad, and such reshaping via pruning might not fit some sparsity patterns, depending on the context.
The hypothesis on the shape can be a relevant limitation for such use cases.
The constructions proposed by \citep{Burkholz22,Burkholz23} appear as a promising direction to overcome this limitation, with the added benefit of reducing the depth overhead.

The convolution operation commonly employed in CNNs can be cumbersome at many points of our analysis.
Exploring different concepts of convolution can be an interesting path for future work as it could lead to tidier proofs and more general results.
For instance, employing a 3D convolution would spare a factor $c$ in \cref{thm:slth_kernel}.

Another limitation of our results is the restriction to ReLU as the activation function.
Many previous works on the SLTH exploit the fact that ReLU satisfies the identity \(x = \relu(x) - \relu(-x)\).
\citep{Burkholz22} leveraged that to obtain an SLTH result for CNNs with activation functions $f$ for which $f(x) - f(-x) \approx x$ around the origin.
Our analysis, on the other hand, does not rely on such property, so adapting the approach of \citep{Burkholz22} to our setting is not straightforward.

Moreover, we remark that the assumption of normally distributed weights might be relaxed.
\citep{borstIntegralityGapBinary2022} provided an MRSSP result for independent random variables whose distribution converges ``fast enough'' to a Gaussian one.\footnote{
  The required convergence rate is higher than that ensured by the Berry-Esseen theorem.
}
We believe our arguments can serve well as baselines to generalise our results to support random weights distributed as such.

Finally, we observe that performing meaningful experiments in this setting is hard.
Directly solving the subset-sum problem with solvers such as Gurobi (as in, e.g., \citep{PensiaRNVP20,daCunhaNV22}) would be prohibitively expensive as the multidimensional version of the RSSP is much harder to solve directly.
An alternative would be to extend an algorithm such as edge pop-up \citep{Ramanujan20} to our setting (structured pruning).
This is a natural and very interesting direction for future work, but it is worth a whole paper on its own.
 
  \section*{Acknowledgments}
  Arthur da Cunha is supported by the European Union (ERC, TUCLA, 101125203).
  Views and opinions expressed are however those of the author(s) only and do not necessarily reflect those of the European Union or the European Research Council.
  Neither the European Union nor the granting authority can be held responsible for them.
  Francesco d'Amore is supported by the Italian Ministry of Universities and Research, project Decreto MUR n. 47/2025, CUP D13C25000750001.
  Emanuele Natale is supported by the French government, through the France 2030 investment plan managed by the Agence Nationale de la Recherche, as part of the "UCA DS4H" project, reference ANR-17-EURE-0004.

  \newpage
  \appendix
\section{Technical tools}\label{app:tools}

\subsection{Concentration inequalities}\label{app:concentration}

\begin{lemma}[Second moment method]
  \label{lemma:secondmoment}
  If $Z$ is a non-negative random variable
  then
  \[
    \Pr\left(Z>0\right) \ge \frac{\expect Z^{2}}{\expect{Z^{2}}}
    .
  \]
\end{lemma}

\begin{lemma}[Chernoff-Hoeffding bounds \citep{dubhashi2011}]\label{lemma:chernoff-hoeffding}
  Let $X_1, X_2, \ldots, X_n$ be independent random variables such that $\Pr\left({0 \le X_i \le 1}\right) = 1$ for all $i \in [n]$.
  Let $X = \sum_{i = 1}^n X_i$ and $\mathbb{E}[{X}] = \mu$.
  Then, for any $\delta \in (0, 1)$ the following holds:
  \begin{enumerate}
    \item if $\mu \le \mu_+$, then
    $
      \Pr\left({X \ge (1 + \delta)\mu_+}\right) \le \exp\left(-\frac{\delta^2 \mu_+}{3}\right);
    $
    \item if $0 \le \mu_-\le \mu$, then
    $
      \Pr\left({X \le (1 - \delta)\mu_-}\right) \le \exp\left(-\frac{\delta^2 \mu_+}{2}\right)
      .
    $
  \end{enumerate}
\end{lemma}

\begin{lemma}[{Corollary of \cite[Lemma 1]{laurentAdaptiveEstimationQuadratic2000}}]
  \label{lemma:chisquared_bound}Let $X\sim\chi_{d}^{2}$ be a chi-squared random variable with
  $d$ degrees of freedom.
  For any \(t > 0\), it holds that
  \begin{enumerate}
    \item \(\Pr\left(X \ge d + 2\sqrt{dt} + 2t\right) \le \exp\left(-t\right)\);
    \item \(\Pr\left(X \le d-2\sqrt{dt}\right) \le \exp\left(-t\right)\)
    .
  \end{enumerate}
\end{lemma}

\begin{lemma}\label{lemma:monotonicity:intersection}
  Let \(X_1, \ldots, X_n, Y_1, \ldots Y_n\) be i.i.d.\ random variables taking values in \(\R\).
  For \(k \in \{0, 1, \ldots, n\}\), define
  \(S_k = \sum_{i = 1}^{n - k}X_i + \sum_{i = n-k + 1}^n Y_i\).
  Then, for all \(k \in \{0, 1, \ldots, n-1\}\), for all \(z \in \R\), and for all \(\eps > 0\), it holds that
  \[
    \pr{S_0 \in \interval{\eps}(z) \cap S_k \in \interval{\eps}(z)} \ge \pr{S_0 \in \interval{\eps}(z) \cap S_{k + 1} \in \interval{\eps}(z)}
    .
  \]
\end{lemma}
\begin{proof}
  Let us proceed by strong induction on \(n\).
  If \(n = 1\) the thesis is trivial.
  We assume \(n > 1\) and the thesis to be true for all values \(1, \ldots, n-1\).
  Notice that \(S_0, S_k, S_{k + 1}\) can be written as follows:
  \begin{align*}
    \begin{cases}
      &S_0 = X_1 + \ldots + X_{n-k-1} + X_{n-k} + X_{n - k + 1} + \ldots + X_n ;
      \\ &S_k = X_1 + \ldots + X_{n-k-1} + X_{n-k} + Y_{n - k + 1} + \ldots + Y_n ;
      \\ &S_{k + 1} = X_1 + \ldots + X_{n-k-1} + Y_{n - k} + Y_{n - k + 1} + \ldots + Y_n
      .
    \end{cases}
  \end{align*}
  Let us define:
  \begin{align*}
    \begin{cases}
      &A_k = X_{n-k} + X_{n - k + 1} + \ldots + X_n ;
      \\ &B_k = X_{n-k} + Y_{n - k + 1} + \ldots + Y_n ;
      \\ &C_k = Y_{n-k} + Y_{n - k + 1} + \ldots + Y_n ;
      \\ &D_k = X_1 + \ldots + X_{n-k-1}
      .
    \end{cases}
  \end{align*}
  Assume \(k \le n-2\).
  The following holds:
  \begin{align}
    & \pr{S_0 \in \interval{\eps}(z), S_k \in \interval{\eps}(z)} \nonumber
    \\ = \ & \expectwrt{X_1, \ldots, X_{n-k-1}}{\pr{A_k \in \interval{\eps}(z - D_k), B_k \in \interval{\eps}(z - D_k) \st X_1, \ldots, X_{n-k-1}}} \nonumber
    \\ \ge \ & \expectwrt{X_1, \ldots, X_{n-k-1}}{\pr{A_k \in \interval{\eps}(z - D_k), C_k \in \interval{\eps}(z - D_k) \st X_1, \ldots, X_{n-k-1}}} \label{eq:intersection:induction}
    \\ = \ & \pr{S_0 \in \interval{\eps}(z), S_{k + 1} \in \interval{\eps}(z)}, \nonumber
    ,
  \end{align}
  where \cref{eq:intersection:induction} holds by the inductive hypothesis.
  Now we just have to address the case \(k = n-1\).
  It holds that
  \begin{align}
    & \pr{S_0 \in \interval{\eps}(z), S_{n-1} \in \interval{\eps}(z)} \nonumber
    \\ = \ & \expectwrt{X_1}{\pr{A_{n-2} \in \interval{\eps}(z - X_1), C_{n-2} \in \interval{\eps}(z - X_1) \st X_1}} \nonumber
    \\ = \ & \expectwrt{X_1}{\pr{A_{n-2} \in \interval{\eps}(z - X_1) \st X_1} \pr{C_{n-2} \in \interval{\eps}(z - X_1) \st X_1}} \label{eq:intersection:independence}
    \\ = \ & \expectwrt{X_1}{\pr{A_{n-2} \in \interval{\eps}(z - X_1) \st X_1}^2} \label{eq:intersection:equally-distributed1}
    \\ \ge \ & \expectwrt{X_1}{\pr{A_{n-2} \in \interval{\eps}(z - X_1) \st X_1}}^2 \label{eq:intersection:jensen}
    \\ = \ & \expectwrt{X_1}{\pr{A_{n-2} \in \interval{\eps}(z - X_1) \st X_1}}\expectwrt{X_1}{\pr{C_{n-2} \in \interval{\eps}(z - X_1) \st X_1}} \label{eq:intersection:equally-distributed2}
    \\ = \ & \pr{S_0 \in \interval{\eps}(z)}\pr{S_{n} \in \interval{\eps}(z)} \nonumber
    ,
  \end{align}
  where \cref{eq:intersection:independence} holds by independence between \(A_{n-2}\) and \(C_{n-2}\), \cref{eq:intersection:equally-distributed1,eq:intersection:equally-distributed2} holds because \(A_{n-2}, C_{n-2}\) are identically distributed, and \cref{eq:intersection:jensen} holds by Jensen's inequality.
\end{proof}

\subsection{Radially-monotone functions}\label{app:radial}

The lemma is expressed in terms of the following definition.
\begin{definition}\label{def:radial}
  We say that a function \(f : \R \to \R\) is \emph{radially-non-increasing} if and only if for any \(x, y \in \R\) with \(\abs{x} \le \abs{y}\) it holds that \(f(x) \ge f(y)\).
\end{definition}
\begin{remark}\label{remark:radial-even}
  Any radially-non-increasing function \(f\) is even as both \(f(x) \le f(-x)\) and \(f(x) \ge f(-x)\) must hold.
\end{remark}

In this section, for any random variable \(\rX\), we denote its density function by \(\varphi_\rX\).

\begin{claim}[Most-probable interval]
  \label{claim:probable_interval}
  Let $\rX$ be a random variable with zero mean.
  If \(\varphi_\rX\) is radially-non-increasing,
  then, for any $z \in \R$ and \(\eps > 0\), we have
  \[
    \pr{\rX \in \left[z-\eps, z + \eps\right]}
    \le \pr{\rX \in \left[-\eps, \eps\right]}
    .
  \]
\end{claim}
\begin{proof}
  We have that,
  \begin{align*}
    \pr{\rX \in \left[-\eps, \eps\right]} - \pr{\rX \in \left[z-\eps, z + \eps\right]}
     = \int_{-\eps}^{\eps} \varphi_\rX(x) \dd{x} - \int_{z-\eps}^{z + \eps} \varphi_\rX(x) \dd{x}
    .
  \end{align*}
  If \(z - \eps \ge \eps \) or \(z + \eps \le -\eps\), the thesis is trivial as \(\varphi_\rX(\abs{x})\) is non-increasing in \(x\).
  W.l.o.g., suppose \(z\) is positive and \(z-\eps < \eps\).
  Then, \(-\eps < z-\eps < \eps < z + \eps\)
  so that
  \begin{align*}
    \int_{-\eps}^{\eps} \varphi_\rX(x) \dd{x} - \int_{z-\eps}^{z + \eps} \varphi_\rX(x) \dd{x}
     & = \int_{-\eps}^{z-\eps} \varphi_\rX(x) \dd{x} - \int_{\eps}^{z + \eps} \varphi_\rX(x) \dd{x}
    \\ & = \int_{-\eps}^{z-\eps} \varphi_\rX(x) - \varphi_\rX({x + 2\eps}) \dd{x}
  \end{align*}
  which is non-negative since \(\varphi_\rX(x) \ge \varphi_\rX({x + 2\eps})\) for \(x \ge -\eps\).
\end{proof}

\begin{claim}\label{claim:app:radial:iff}
  Assume \(f\colon \R \to \R\) to be a radially-non-increasing differentiable function.
  Then, \(f'\) is an odd function which is non-positive for \(x \ge 0\) and non-negative for \(x \le 0\).
\end{claim}
\begin{proof}
  We have that \(f\) is even if and only if \(f'\) is odd.
  Moreover, \(f\) is non-increasing for \(x \ge 0\) if and only if \(f'(x)\) is non-positive and,
  analogously, \(f\) is non-decreasing for \(x \le 0\) if and only if \(f'(x)\) is non-negative.
\end{proof}

Recall that a function \(f\colon \R \to \R\) is absolutely integrable if there exists \(M \in \R\) such that \(\int_{\R} \abs{f(x)} \dd{x} = M\).

\begin{lemma}\label{lemma:app:radial-sum}
  Let \(\rX, \rY\) be independent random variables with radially-non-increasing PDFs.
  If the derivative of \(\varphi_\rX\) is well-defined everywhere and is absolutely integrable,
  then \(\varphi_{\rX + \rY}\) is also radially-non-increasing.
\end{lemma}
\begin{proof}
  With \cref{claim:app:radial:iff} in mind,
  we analyze the derivative of \(\varphi_{\rX + \rY}\).

  We have that \(\varphi_{\rX + \rY}(z) = (\varphi_\rX \circ \varphi_\rY) (z)\).
  Since by hypothesis both \(\varphi_\rX\) is differentiable and PDFs are always absolutely integrable,
  \(\varphi_{\rX + \rY}'(z) = (\varphi_\rX' \circ \varphi_\rY) (z)\).
  Given \(z \ge 0\),
  to see that \(\varphi_{\rX + \rY}'(z) \le 0\)
  notice that
  \begin{align*}
    (\varphi_\rX' \circ \varphi_\rY)(z)
     & = \int_{\R} \varphi_\rX'(x) \varphi_\rY(z - x) \dd{x}
    \\ & = \int_{-\infty}^{0} \varphi_\rX'(x) \varphi_\rY(z - x) \dd{x} + \int_{0}^{+\infty} \varphi_\rX'(x) \varphi_\rY(z - x) \dd{x}
    \\ & = \int_{-\infty}^{0} -\varphi_\rX'(-x) \varphi_\rY(z - x) \dd{x} + \int_{0}^{+\infty} \varphi_\rX'(x) \varphi_\rY(z - x) \dd{x} \tag{as \(\varphi_\rX\) is even}
    \\ & = \int_{0}^{+\infty} -\varphi_\rX'(x) \varphi_\rY(z + x) \dd{x} + \int_{0}^{+\infty} \varphi_\rX'(x) \varphi_\rY(z - x) \dd{x} \tag{by substitution (\(x' = -x\))}
    \\ & = \int_{0}^{+\infty} \varphi_\rX'(x) \left[\varphi_\rY(z - x) - \varphi_\rY(z + x)\right] \dd{x}
    .
  \end{align*}
  For \(x \ge 0\) it holds both that \(\varphi_\rX'(x) \le 0\).
  Since we are also assuming that \(z \ge 0\),
  we obtain that \(\abs{z - x} \le \abs{z + x}\)
  so that \(\varphi_\rY(z - x) - \varphi_\rY(z + x) \ge 0\)
  by the radial-non-increase of \(\varphi_\rY\).
  Hence, \((\varphi_\rX' \circ \varphi_\rY) (z) \le 0\).
  The symmetrical argument proves that \((\varphi_\rX' \circ \varphi_\rY) (z) \ge 0\) for \(z \le 0\).

  Finally,
  to see that \(\varphi_\rX' \circ \varphi_\rY\) is odd,
  notice that
  \begin{align*}
    (\varphi_\rX' \circ \varphi_\rY) (-z)
     & = \int_{\R} \varphi_\rX'(x) \varphi_\rY(-z - x) \dd{x}
    \\ & = \int_{\R} \varphi_\rX'(x) \varphi_\rY(z + x) \dd{x} \tag{as \(\varphi_\rY\) is even}
    \\ & = \int_{\R} \varphi_\rX'(-x) \varphi_\rY(z - x) \dd{x} \tag{by substitution (\(x' = -x\))}
    \\ & = \int_{\R} - \varphi_\rX'(x) \varphi_\rY(z - x) \dd{x} \tag{as \(\varphi_\rX'\) is odd}
    \\ & = - (\varphi_\rX' \circ \varphi_\rY) (z)
    .
  \end{align*}
\end{proof}

\begin{claim}\label{claim:app:radial:product}
  If \(f \colon \R \to \R\) and \(g \colon \R \to \R\) are non-negative radially-non-increasing functions,
  then \(f \cdot g\) is also non-negative and radially-non-increasing.
\end{claim}
\begin{proof}
  Since \(f\) and \(g\) are non-negative, so is \(f \cdot g\).
  For the rest,
  let \(z_1, z_2 \in \R \) with \(0 \le \abs{z_1} \le \abs{z_2}\)
  so that
  \begin{align*}
    (f \cdot g)(z_1) & = f(z_1) \cdot g(z_1)
    \\ & = f(\abs{z_1}) \cdot g(\abs{z_1}) \tag{by \cref{remark:radial-even}}
    \\ & \ge f(\abs{z_2}) \cdot g(\abs{z_2}) \tag{by radial-non-increase and non-negativity}
    \\ & = f(z_2) \cdot g(z_2) \tag{by \cref{remark:radial-even}}
    \\ & = (f \cdot g)(z_2)
    .
  \end{align*}
\end{proof}

\begin{lemma}\label{lemma:tool:integral1-MRSSP:clever}
  Let \(\rA, \rB, \rC\) be independent random variables with zero mean and radially-non-increasing PDFs.
  If the derivatives of \(\varphi_\rA\) and \(\varphi_\rB\) exist everywhere and are absolutely integrable,
  then for all \(z \in \R\) and \(\eps > 0\) it holds that
  \[
    \pr{\rA + \rB \in (z - \eps, z + \eps), \rA + \rC \in (z - \eps, z + \eps)} \le \pr{\rA + \rB \in (-\eps, + \eps), \rA + \rC \in (-\eps, + \eps)}
    .
  \]
\end{lemma}
\begin{proof}
  We have that
  \begin{align*}
    & \pr{\rA + \rB \in (z - \eps, z + \eps), \rA + \rC \in (z - \eps, z + \eps)}
    \\ = \ & \int_\R \varphi_\rB(x) \left[\int_{z - x - \eps}^{z - x + \eps} \varphi_\rA(y) \dd{y} \cdot \int_{z - x - \eps}^{z - x + \eps} \varphi_\rC(y) \dd{y} \right] \dd{x}
    \\ = \ & \int_\R \varphi_\rB(z - s) \left[\int_{s - \eps}^{s + \eps} \varphi_\rA(y) \dd{y} \cdot \int_{s - \eps}^{s + \eps} \varphi_\rC(y) \dd{y} \right] \dd{s}
    ,
  \end{align*}
  where we substituted \(x = z - s\).
  Letting \(\rU \sim \Uniform([-\eps, + \eps])\),
  we obtain that
  \begin{align*}
    \int_{s - \eps}^{s + \eps} \varphi_\rA(y) \dd{y}
     & = \int_{-\eps}^{+\eps} \varphi_\rA(s - y) \dd{y}
    \\ & = 2\eps \cdot \int_{-\eps}^{+\eps} \frac{1}{2\eps}\varphi_\rA(s - y) \dd{y}
    \\ & = 2\eps \cdot \int_{-\eps}^{+\eps} \varphi_\rU(s) \varphi_\rA(s - y) \dd{y}
    \\ & = 2\eps \cdot \varphi_{\rA + \rU}(s)
    .
  \end{align*}
  Similarly, \(\int_{s - \eps}^{s + \eps} \varphi_\rC(y) \dd{y} = 2\eps \cdot \varphi_{\rC + \rU}\),
  hence
  \begin{align}
    & \pr{\rA + \rB \in (z - \eps, z + \eps), \rA + \rC \in (z - \eps, z + \eps)} \nonumber \\
    = \ & 4\eps^2 \int_\R \varphi_\rB(z - x) \cdot \varphi_{\rA + \rU}(x) \cdot \varphi_{\rC + \rU}(x) \dd{x} \label{eq:radial:clever:1}
    .
  \end{align}
  Now consider the function \(f = \varphi_{\rA + \rU} \cdot \varphi_{\rC + \rU}\).
  We proceed to show that \(f\) is radially-non-increasing and, up to a normalization factor, the PDF of some random variable.
  Indeed,
  by \cref{lemma:app:radial-sum} both \(\varphi_{\rA + \rU}\) and \(\varphi_{\rC + \rU}\) are non-negative and radially-non-increasing
  thus, by \cref{claim:app:radial:product}, so is their product, \(f\).
  To see that \(f\) is also absolutely integrable,
  notice that since the radial-non-increase of \(\varphi_\rA\) implies that \(\varphi_\rA(x) \le \varphi_\rA(0) \eqqcolon M\) for all $x \in \R$,
  we must have that \(\abs{\varphi_{\rA + \rU}(x)} = \varphi_{\rA + \rU}(x) \le 2\eps \sup_{y \in (x-\eps, x + \eps)} \varphi_\rA(x) \le 2\eps M\).
  Then,
  \begin{align*}
    \int_{\R} f(x) \dd{x}
     & = \int_{\R} \abs{f(x)} \dd{x} \tag{as \(f\) is non-negative}
    \\ & = \int_{\R} \abs{\varphi_{\rA + \rU}(x) \cdot \varphi_{\rC + \rU}(x)} \dd{x}
    \\ & \le \sup_{x \in \R} \Set*{\abs{\varphi_{\rA + \rU}(x)}} \cdot \int_{\R} \abs{\varphi_{\rC + \rU}(x)} \dd{x} \tag{by Hölder's inequality}
    \\ & \le 2\eps M \cdot 1 \tag{as \(\varphi_{\rC + \rU}\) is a PDF}
    ,
  \end{align*}
  where for the last inequalities we also used our previous observation
  and for Hölder's inequality we employed the pair \((\infty, 1)\) of Hölder conjugates.

  Overall,
  we conclude that there must exist random variable \(\rD\) with radially-non-increasing PDF given by \(\Par*{\int_{\R} f(x) \dd{x}}^{-1} \cdot f\).
  Applying this to \cref{eq:radial:clever:1},
  we obtain that
  \begin{align*}
    & \pr{\rA + \rB \in (z - \eps, z + \eps), \rA + \rC \in (z - \eps, z + \eps)} \\
     = \ & \frac{4\eps^2}{\int_\R \varphi_{\rA + \rU}(x) \cdot \varphi_{\rC + \rU}(x) \dd{x} } \int_\R \varphi_\rB(z - x) \cdot \varphi_{\rD}(x) \dd{x}
    \\ = \ & \frac{4\eps^2}{\int_\R \varphi_{\rA + \rU}(x) \cdot \varphi_{\rC + \rU}(x) \dd{x}} \cdot (\varphi_\rB \ast \varphi_\rD)(z)
    \\ = \ & \frac{4\eps^2}{\int_\R \varphi_{\rA + \rU}(x) \cdot \varphi_{\rC + \rU}(x) \dd{x}} \cdot \varphi_{\rB + \rD}(z)
    .
  \end{align*}
  Finally,
  by \cref{lemma:app:radial-sum} we have that \(\varphi_{\rB + \rD}(z)\) is radially-non-increasing as well, so that it reaches its maximum at \(z = 0\), which implies the thesis.
\end{proof}

\subsection{Results from \texorpdfstring{\citep{becchettiMultidimensionalRandomSubset2022}}{becchettiMultidimensionalRandomSubset2022}}\label{app:mrrs-from-arxiv}

The following result is a technical estimation provided in \citep{becchettiMultidimensionalRandomSubset2022}.

\begin{lemma}\label{lemma:tool:integral2-MRSSP}
  For all \(x \in \R\), \(c \in (0, 1/162)\), and \(\eps \in (0, 1)\), it holds that
  \[
    \left(\int_{-\eps}^{+\eps} e^{-c(x + s)^2} \dd{s}\right) ^2 \le \left(\int_{-\eps}^{+\eps} \frac{e^{-c(x + \eps)^2} + e^{-c(x-\eps)^2}}{2} \cdot e^{c\eps^2} \dd{s}\right)^2
    .
  \]
\end{lemma}

\subsection{Supporting results}\label{app:supporting-results}

\begin{lemma}\label{lemma:find-maximum:2}
  Let \(a \in (0, a_0)\) be a real number.
  Consider the function
  \begin{align*}
    f(x) = \frac{\exp\left[-\frac{a}{b + y}\right]}{\sqrt{b + x}}
  \end{align*}
  for \(x \in [x_0, + \infty)\) and \(b \ge a_0\), with \(x_0 \in [0, + \infty)\).
  Then, the global maximum of the function is in \(x = x_0\).
\end{lemma}
\begin{proof}
  Notice that \(f(x)\) is continuous and differentiable.
  Let us compute \(\frac{\dd {f(x)}}{\dd x}\).
  \begin{align*}
    \frac{\dd {f(x)}}{\dd x} & = \frac{\exp \left[-\frac{a}{b + x}\right] \cdot \frac{a}{(b + x)^2} \cdot \sqrt{b + x} - \exp\left[-\frac{a}{b + x}\right] \cdot \left[\frac{1}{2\sqrt{b + x}}\right]}{b + x}
    \\ & = \frac{\exp \left[-\frac{a}{b + x}\right] \cdot \left[2a -b -x\right]}{2(b + x)^2\sqrt{b + x}}
    .
  \end{align*}
  The minimum in \(x\) is reached in \(x = 2a - b \le 0\) and the function is non-increasing for \(x \ge 0\).
  Hence, the global minimum is reached in \(x = x_0\).
\end{proof}

\begin{lemma}[NSN with positive scalar]
  \label{lemma:halfnormalNSN} If a $d$-dimensional random vector $Y$
  is such that, for each $i \in \left[d\right]$, $Y_{i} = \widetilde{Z} \cdot \widetilde{Z}_{i}$, where \(\widetilde{Z}_{1}, \ldots, \widetilde{Z}_{n}\) are identically distributed random variables following
  a standard normal distribution, $\widetilde{Z}$ is a half-normal distribution,\footnote{I.e. $\widetilde{Z} = \abs{Z}$ where $Z$ is a standard normal distribution.} and
  $\widetilde{Z}, \widetilde{Z}_{1}, \ldots, \widetilde{Z}_{n}$ are independent,
  then $Y$ follows an NSN distribution.
\end{lemma}

\begin{proof}
  By \cref{def:nsn_vector}, $Y$ is NSN if, for each $i \in \left[d\right]$,
  $Y_{i} = Z \cdot Z_{i}$ where $Z, Z_{1}, \ldots, Z_{n}$ are i.i.d.\ random
  variables following a standard normal distribution.
  We know that \(\widetilde{Z} = \abs{Z}\), therefore \(\widetilde{Z}_i = \sign{Z}\sign{Z_i}\abs{Z_i}\) for each \(i = 1, \ldots, n\), where \(Z, Z_1, \ldots, Z_n\) are i.i.d.\ standard normal random variables, as \(\sign{Z}\sign{Z_i}\) is independent of \(\sign{Z}\) and of \(\sign{Z}\sign{Z_j}\) for \(i \neq j\).
  Then,
  \begin{align*}
    Y_{i}
     & = \widetilde{Z} \cdot \widetilde{Z}_{i}
    \\ & = \abs{Z} \cdot \sign{Z}\sign{Z_i}\abs{Z_i}
    \\ & = \sign{Z}\abs{Z} \cdot \sign{Z_i}\abs{Z_i}
    \\ & = Z \cdot Z_i
    ,
  \end{align*}
  implying the thesis.
\end{proof}

\begin{corollary}[of \cref{thm:mrrs_gaussian_scaled}]
  \label{cor:mrrs_NSN}
  Let \(0 < \eps < 1/4\).
  Let \(d\) and \(n\) be positive integers and consider \(n = \Ceil{C d^4(\log \frac 1\eps + \log d)}\) as given by \cref{thm:mrrs_gaussian_scaled}.
  Consider \(N\) \(d\)-dimensional i.i.d.\ NSN random vectors $X_{1}, \ldots, X_{N}$ with \(N \ge n\).
  There exist a universal constant \(C'\) such that, if \(N \ge C' d \log \frac{d}{\eps}\) and \(k = n/(6\sqrt{d})\),
  it holds that
  \[
    \Pr\left(\forall
    \target \in \R^{d}: \normone*{\target} \le \sqrt{k}, \exists S, \abs{S} = k : \normmax*{\left(\sum_{i \in S}X_{i}\right)-\target} \le \eps\right) \ge 1-\eps
    .
  \]
  \end{corollary}

\begin{proof}
  Observe that the set $[-\sqrt{k}, \sqrt{k}]^d$ can be ``partitioned'' in $(\sqrt{k}/\eps)^d$ many \(\infty\)-norm balls of radius $\eps$, which we denote by \(B_1, \ldots, B_{(\sqrt{k}/\eps)^d}\).
  Notice that \(\normone{\target} \le \sqrt{k} \implies \target \in [-\sqrt{k}, \sqrt{k}]^d\).
  Let $s$ be a positive value to be fixed later with the constraint that \(n s \le N\).
  Let us partition the $N$ vectors $X_{1}, \ldots, X_{N}$ in $s$ disjoint sets
  $G_{1}, \ldots, G_{s}$ of at least $n$ vectors each.
  By \cref{thm:mrrs_gaussian_scaled},
  there is a constant $ \delta \in \left(0, 1\right)$ such that for each group
  $G_{i}$ ($i \in \left[s\right]$), and for any target \(\target\) such that \(\normone{\target} \le \sqrt{k}\),
  \begin{align}
    \Pr\left(\exists S\subseteq G_{i}, \abs{S} = k : \normmax*{\left(\sum_{i \in S}X_{i}\right)-\target} \le \eps\right) \ge \delta.\nonumber
  \end{align}
  Notice that the above probability is independent of \(i \in [s]\) since the normal vectors \(X_i\) are i.i.d.\ and the groups identified by the \(G_i\)s are disjoint.
  Hence, by independence,
  \begin{align}
    \Pr\left(\exists i \in [s] : \exists S\subseteq G_{i}, \abs{S} = k : \normmax*{\left(\sum_{i \in S}X_{i}\right)-\target} \le \eps\right) \ge 1 - (1-\delta)^s
    .\label{eq:prob_all_groups}
  \end{align}
  Let \(\target_\star\) be a target vector such that \(\normone{\target_\star} \le \sqrt{k}\) and such that \cref{eq:prob_all_groups} is minimized.
  It follows that
  \begin{align}
    & \Pr\left(\forall
    \target \in \R^{d}, \normone{\target} \le \sqrt{k}, \exists S \subseteq [N], \abs{S} = k : \normmax*{\left(\sum_{i \in S}X_{i}\right)-\target} \le \eps\right) \nonumber
    \\ \ge \ & \Pr\left(\forall
    \target \in \R^{d}, \normone{\target} \le \sqrt{k}, \exists i \in \left[s\right], \exists S\subset G_{i}, \abs{S} = k : \normmax*{\left(\sum_{i \in S}X_{i}\right)-\target} \le \eps\right) \nonumber
    \\ = \ & 1-\Pr\left(\exists
    \target \in \R^{d}, \normone{\target} \le \sqrt{k}, \forall i \in \left[s\right], \forall S\subset G_{i}, \abs{S} = k : \normmax*{\left(\sum_{i \in S}X_{i}\right)-\target} > \eps\right)\nonumber
    \\ \ge \ & 1 - \left(\frac{\sqrt{k}}{\eps}\right)^d \Pr\left(\forall i \in [s], \forall S \subset G_{i}, \abs{S} = k : \normmax*{\left(\sum_{i \in S}X_{i}\right)-\target_\star} > \eps\right) \label{eq:boost:ub}
    \\ \ge \ & 1-\left(\frac{\sqrt{k}}{\eps}\right)^d \left(1-\delta\right)^s
    ,\label{eq:boost:indep}
  \end{align}
  where \cref{eq:boost:ub} holds by the union bound over all \(B_i\)s, \cref{eq:boost:indep} comes from \cref{eq:prob_all_groups}.
  If we choose \(s = \Floor*{C' d \log \frac{\sqrt{k}}{\eps}}\) with \(C'\) large enough, we obtain
  \[
    1-(k/\eps)^{d}\left(1-\delta\right)^{\Floor*{C' d \log \frac{\sqrt{k}}{\eps}}} = 1-\exp\left[{d \log \frac{1}{\eps} - \Floor*{C' d \log \frac{\sqrt{k}}{\eps}} \log \frac{1}{1 - \delta} } \right] \ge 1 - \eps
    .
  \]
  Since \(k = n / (6\sqrt{d})\), it is sufficient to take \(N \ge C'' dn \log \frac{d}{\eps}\) to ensure that \(n s \le N\) for a large enough constant \(C''\).
\end{proof}

\begin{lemma}[Tensor Convolution Inequality]
  \label{lemma:convineq}
  Given real tensors $\tK$ and $\tX$ of respective
  sizes $d\times d^{\prime}\times c_{0}\times c_{1}$ and $D\times D^{\prime}\times c_{0}$,
  it holds that
  \[
    \normmax*{\tK\ast \tX} \le \normone* \tK \cdot \normmax* \tX
    .
  \]
\end{lemma}

\begin{proof}
  We have
  \begin{align*}
    \normmax*{\tK\ast \tX}
     & \le \max_{i, j \in \left[D\right], \ell \in \left[c_{1}\right]}\sum_{i^{\prime}, j^{\prime} \in \left[d\right], k \in \left[c\right]}\left|\tK_{i^{\prime}, j^{\prime}, k, \ell}X_{i-i^{\prime} + 1, j-j^{\prime} + 1, k}\right|
    \\ & \le \max_{i, j \in \left[D\right], \ell \in \left[c_{1}\right]}\left(\sum_{i^{\prime}, j^{\prime} \in \left[d\right], k \in \left[c\right]}\left|\tK_{i^{\prime}, j^{\prime}, k, \ell}\right|\right)\normmax* \tX
    \\ & \le \max_{i, j \in \left[D\right], \ell \in \left[c_{1}\right]}\normone* \tK \cdot \normmax* \tX
    \\ & = \normone* \tK \cdot \normmax* \tX
    .
  \end{align*}
\end{proof}

  \bibliographystyle{plainnat}
   \vskip 0.2in
  \bibliography{biblio}
\end{document}